\documentclass{article}

\usepackage{fullpage}
\usepackage{amssymb}
\usepackage{amsfonts}
\usepackage{mathrsfs}
\usepackage{amsthm}
\usepackage{amsmath}
\usepackage{mathtools}
\usepackage{caption}
\usepackage{subcaption}
\usepackage{multirow}
\usepackage{graphicx}
\usepackage{booktabs}
\usepackage[export]{adjustbox}
\usepackage{pgfplots}
\pgfplotsset{compat=1.15}

\newtheorem{observation}{Observation}
\newtheorem{proposition}{Proposition}

\DeclareMathOperator{\cut}{\operatorname{Cut}}
\DeclareMathOperator{\ncut}{\operatorname{NCut}}

\DeclareMathOperator{\mixcut}{\operatorname{MixCut}}
\DeclareMathOperator{\mixncut}{\operatorname{MixNCut}}

\DeclareMathOperator{\vol}{\operatorname{Vol}}
\DeclareMathOperator{\length}{\operatorname{Len}}

\DeclareMathOperator*{\jac}{\operatorname{Jaccard}}

\begin{document}

\title{Spectral Image Segmentation with Global Appearance Modeling}

\author{%
  Jeova F. S. Rocha Neto\thanks{Department of Computer Science,
  Haverford College, Haverford, PA, USA}
  \and Pedro F. Felzenszwalb\thanks{School of Engineering, Brown
    University, Providence, RI, USA}
  }
\maketitle

\begin{abstract}
We introduce a new spectral method for image segmentation that
incorporates long range relationships for global appearance
modeling. The approach combines two different graphs, one is a sparse
graph that captures spatial relationships between nearby pixels and
another is a dense graph that captures pairwise similarity between
\emph{all pairs of pixels}. We extend the spectral method for
Normalized Cuts to this setting by combining the transition matrices
of Markov chains associated with each graph.  We also derive an
efficient method for sparsifying the dense graph of appearance
relationships.  This leads to a practical algorithm for segmenting
high-resolution images.  The resulting method can segment challenging
images without any filtering or pre-processing.
\end{abstract}
 
\section{Introduction}

Image segmentation is a fundamental problem in computer vision.
Spectral clustering methods pioneered by the normalized cuts approach
\cite{shi2000normalized} provide simple and powerful algorithms based
on fundamental graph-theoretic notions and computational linear
algebra.

\begin{figure}
\centering
    \begin{subfigure}[t]{0.29\textwidth}
        \includegraphics[width = \textwidth, height = 2.4cm]{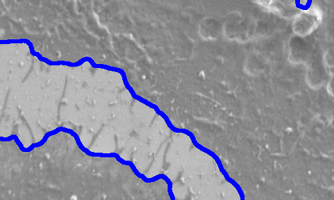}
    \end{subfigure}
    \hspace{1pt}
    \begin{subfigure}[t]{0.17\textwidth}
        \includegraphics[width = \textwidth, height = 2.4cm]{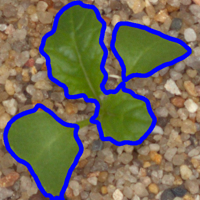}
    \end{subfigure}
    \hspace{1pt}
    \begin{subfigure}[t]{0.23\textwidth}
        \includegraphics[width = \textwidth, height = 2.4cm]{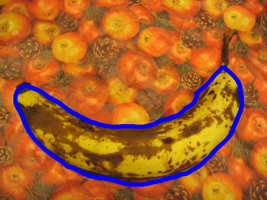}
    \end{subfigure}
    \hspace{1pt}
    \begin{subfigure}[t]{0.26\textwidth}
        \includegraphics[width = \textwidth, height = 2.4cm]{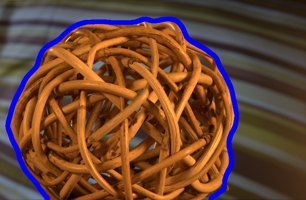}
    \end{subfigure}
    \caption{Segmentation results using the proposed method}
    \label{fig:fig_intro}
\end{figure}

Spectral clustering methods are formulated using an objective function
defined by a graph.  The classical constructions used for image
segmentation focus on pairwise similarity between nearby pixels.  In
this paper we introduce a new spectral method that incorporates long
range relationships for global appearance modeling.  The resulting
method can segment challenging images without any filtering or
pre-processing.  Figure~\ref{fig:fig_intro} shows several results
obtained with the proposed method.  Figure~\ref{fig:real_exp} shows
how the new method significantly outperforms the original normalized
cuts formulation.

We use a dense graph to capture the global appearance of regions.  We show the
normalized cut criteria in this graph reflects the distributions of pixel
values in each region using a kernel density estimate.  The measure
penalizes the overlap between distributions in different regions.

To implement our image segmentation approach we extend the normalized
cuts spectral algorithm to a setting where there are multiple graphs
that encode different grouping cues.  Our approach for image
segmentation combines two graphs.  We provide a natural interpretation
for the normalized cut criteria on each of these graphs.  One of the
graphs is sparse and does not depend on the image data, it simply
captures spatial relationships between pixels.  The other graph is
dense and captures pairwise similarity between \emph{all pairs of
pixels}, irrespective of their spatial location.

The direct implementation of spectral methods to segment high
resolution images is challenging due to high memory and computational
requirements.  We tackle this challenge using a graph sparsification
approach that enables the efficient segmentation of high resolution
images.

We show experimental results with a variety of images and provide a
quantitative evaluation using a dataset of synthetic images with
Brodatz textures.  Our approach achieves highly accurate results in
this setting despite the complex appearance of the textures.

\section{Previous work}

The mathematical foundation of spectral graph partitioning can be
dated back to \cite{donath1973lower} and
\cite{fiedler1973algebraic}. The seminal work of Shi and Malik
\cite{shi2000normalized} built on this foundation to develop the
normalized cuts method for image segmentation. Since then a
significant body of work has emerged in the fields of both spectral
clustering and image segmentation (see,
i.e.\ \cite{cheung2018graph}). An important development on the
theoretical understanding of normalized cuts can be found in
\cite{meila2001learning}, where the authors traced the parallel
between the normalized cuts criteria and low conductivity sets in
Markov chains.

On the application and implementation side, the normalized cuts
algorithm poses two main challenges when applied to image
segmentation: how to (efficiently) construct a graph over the set of
pixels and how to perform the spectral decomposition. In order to
solve them, the usual approach is to construct sparse graphs where
each pixel is connected to nearby pixels \cite{shi2000normalized,
  maji2011biased, chew2015semi}. Furthermore, works such as
\cite{perona1998factorization, fowlkes2004spectral, lin2010power}
reduce the computation burden of spectral segmentation by relying on
approximate eigenvector solvers. Our method instead starts with a
dense graph connecting all pairs of pixels and uses an edge
sparsification technique based on importance sampling.  We also
provide an importance sampling implementation of the original
similarity graph proposed in \cite{shi2000normalized}.

A key step in developing spectral segmentation algorithms is to decide
which information to extract from the image in order to compute pixel
affinities (the edge weights). While the initial implementations used
raw intensities or filter-banks \cite{shi2000normalized,
  meila2001learning, malik2001contour}, further developments adopted
the intervening contour cue from probability of boundary (Pb) maps
\cite{arbelaez2010contour, maji2011biased, chew2015semi}. In this
paper, we show that using raw pixel intensities (or color) is enough
to obtain satisfactory segmentation results in complex images.  In
particular, our method is able to outperform traditional filter-bank
based methods in segmentation of textured images.

Our spectral segmentation algorithm is also related to multiview
clustering methods. In multiview clustering one is concerned on
clustering the data using different features (or views). One of the
earlier results on this problem can be traced to
\cite{bickel2004multi}, where the authors propose multiview
counterparts of traditional clustering algorithms, such as EM and
$K$-means.

In the spectral multiview realm, the work presented in
\cite{de2005spectral} proposes a two-view clustering algorithm that
computes the normalized eigenvectors arising from a bipartite graph
that encodes the views. The authors in \cite{kumar2011co} describe an
iterative procedure that enforces consistency among views by solving
the generalized eigenvalue problem for each view separately using the
eigenvectors from other views computed in previous
iterations. Finally, other relevant approaches to multiview clustering
involve combining graphs and weights arising from different views into
a sole graph either via a convex combination \cite{zhou2007spectral,
  zong2018weighted} or according to their Laplacian's power mean
\cite{mercado18power}. To the best of our knowledge, no multiview
clustering algorithm has been developed for image segmentation. One
possible explanation for this is the computational burden of existing
multi-view clustering methods.

Considering image segmentation broadly speaking, there is an
increasing popularity of deep learning based solutions, some of which
find their inspiration on normalized cuts \cite{xia2017w,
  tang2018regularized}. We refer the interested reader to the work in
\cite{ghosh2019understanding} for a comprehensive survey on these
techniques. Departing from this learning based paradigm, our method
does not require training data and can be applied in different
settings with little or no fine-tuning.  Moreover, our method produces
interpretable results. Furthermore, the proposed algorithm
demonstrates the power and practical use of long range relationships
between pixels, which are not, or at least not explicitly, present
within typical deep learning frameworks.

\section{Background} \label{sec:background}

\subsection{Graph cuts and spectral clustering}

Let $G = (V, E, w)$ be an undirected weighted graph.  A cut $(A,B)$ is
a partition of $V$ into two disjoint sets.  We will consider the weight of
a cut in different graphs that have the same set of vertices.  Let $w(i,j)
= 0$ when $\{i,j\} \not\in E$.  The weight of a cut $(A, B)$ in $G$ is
defined as,
\begin{equation*}
    \cut (A, B|G) = \sum_{i \in A, j \in B} w(i, j).
\end{equation*}

In the context of clustering and image segmentation it is typical to
use large weights to indicate that elements are similar and should not be separated.  In this case we can look for the \emph{minimum cut}
to find an optimal partition of $V$.  However, this strategy is
heavily biased towards imbalanced cuts, such as having a single node
on one side.  This motivated the introduction of the celebrated
normalized cut criteria and algorithm \cite{shi2000normalized}.

The \emph{normalized cut} value is defined as,
\begin{equation*}
    \ncut (A, B|G) = \frac{\cut(A, B|G)}{\vol(A|G)} + \frac{\cut(A, B|G)}{\vol(B| G)} = \vol(V|G)\frac{\cut(A, B|G)}{\vol(A|G)\vol(B| G)}.
\end{equation*}
Here $\vol(A|G)$ is a measure of the ``volume'' of $A$ defined as
$\vol(A|G) = \sum_{i \in A, j \in V} w(i, j)$.  

The spectral algorithm
introduced in \cite{shi2000normalized} solves a continuous relaxation
of the minimum $\ncut$ problem.  Let $W$ be the weighted adjacency
matrix of $G$.  Let $D$ be the diagonal degree matrix with $D(i,i) =
\sum_{j \in V} W(i,j)$.  The matrix $L=D-W$ is the \emph{Laplacian} of
$G$.

The $\ncut$ algorithm solves a generalized eigenvector problem,
\begin{equation}\label{eqn:geigenvector}
Lx = \lambda Dx.
\end{equation}
The algorithm selects the eigenvector $x$ with second smallest
eigenvector, and partitions $V$ by thresholding $x$. In
\cite{meila2001learning} the $\ncut$ criteria and algorithm is
described in terms of a Markov chain.  Let $P=D^{-1}W$.  The matrix
$P$ is the transition matrix of a Markov chain over the vertices $V$.
The long term behavior of this Markov chain can be characterized by
the solutions of the eigenvector problem,
\begin{equation}\label{eqn:eigenvector}
Px = \lambda x.  
\end{equation}
A solution $(\lambda,x)$ to the eigenvector problem in
(\ref{eqn:eigenvector}) leads to a solution $(1-\lambda,x)$ to the
generalized eigenvector problem in (\ref{eqn:geigenvector}) and
vice-versa.  Therefore the generalized eigenvector $x$ used in the
$\ncut$ algorithm corresponds to the eigenvector of $P$ with second
largest eigenvalue.

\subsection{Traditional Graph Construction for Image Segmentation}
\label{sec:original}

The classical application of normalized cuts for image segmentation
involves a graph $H$ where the vertices represent the image pixels and
the weights reflect simultaneously the appearance similarity and
distance between pairs pixels.

Let $H=(V,E,w)$ be a graph where the vertices $V$ are the pixels in an
image and the edges $E$ connect every pair of pixels.  We use $I(j)$
and to denote the appearance (such as the brightness or color) of
pixel $j$.  We use $X(j)$ to denote the spatial location of the same
pixel.  Now define,
\begin{equation}\label{eq:trad_ncut_seg_eq}
 w(i,j) = 
\exp\left(-\frac{||I(i) - I(j)||^2}{2\sigma_I^2}\right) 
\exp\left(-\frac{||X(i) - X(j)||^2}{2\sigma_X^2}\right).
\end{equation}

The graph $H$ combines two grouping cues in a single real valued
weight\footnote{The graph defined here differs slightly from the one
used in \cite{shi2000normalized} because in \cite{shi2000normalized}
the weight of an edge is set to 0 if the distance between $i$ and $j$
is above a threshold.} and shares similarities with bilateral
filtering \cite{tomasi1998bilateral}. Using the normalized cut
criteria, pixels are encouraged to be grouped together if they have
similar appearance \emph{and} are close to each other.  Note, however,
that pixels that have similar appearance but are far away are not
encouraged to be grouped because the corresponding weight is close to
zero.  Similarly, neighboring pixels that have very different
appearance, such as in a textured region, are also not encouraged to
be grouped.

\section{New Criteria for Image Segmentation}
\label{sec:seg_inter}

\begin{figure}[t]
    \centering
    \begin{subfigure}[t]{0.4\textwidth}
        \centering
        \includegraphics[width =.8\textwidth]{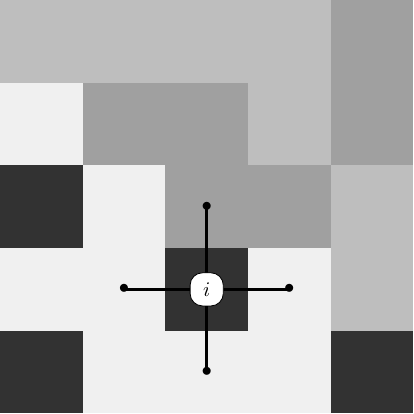}

        \caption{$G_{\text{grid}}$}
    \end{subfigure}
    \begin{subfigure}[t]{0.4\textwidth}
        \centering
        \includegraphics[width =.8\textwidth]{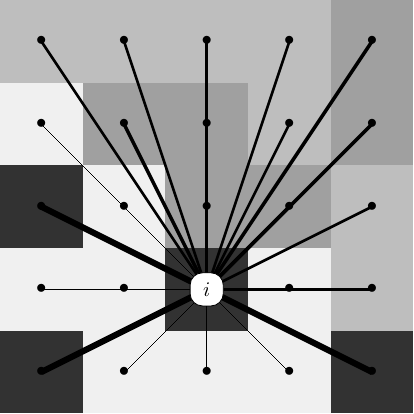}

        \caption{$G_{\text{data}}$}
    \end{subfigure}
    \caption{The edges connecting to a pixel $i$ in $G_{\text{grid}}$
      and $G_{\text{data}}$. In the each image, the thickness of each
      link represents the weight of the edge connecting a pair of
      pixels.}
    \label{fig:graph_imgs}
\end{figure}

We combine two normalized cut values to obtain a new criteria for
image segmentation.  We break the grouping cues (spatial proximity and
appearance similarity) into two separate graphs, $G_{\text{grid}}$ and
$G_{\text{data}}$.  Both graphs are defined over the same set of
vertices, corresponding to the pixels in an image.
\begin{enumerate}
\item The graph $G_{\text{grid}}$ is a grid over the image pixels,
  where each pixel is connected to the four neighboring pixels with an
  edge of weight 1.  This graph encourages neighboring pixels to be
  grouped together, independent of their appearance.
\item The graph $G_{\text{data}}$ is a fully connected graph that
  encourages pixels with similar appearance to be grouped together,
  independent of their location.  The weights in $G_{\text{data}}$ are
  based on appearance similarity of pixels, and do not depend on pixel
  locations,
\begin{equation*}
    w(i, j) = \exp\left(-\frac{||I(i) - I(j)||^2}
    {2\sigma^2}\right).
\end{equation*}
\end{enumerate}
Figure \ref{fig:graph_imgs} illustrates the graph construction of both
$G_{\text{grid}}$ and $G_{\text{data}}$ on a image with two regions.

\subsection{Spatial Information: $G_{\text{grid}}$}

Let $(A,B)$ be a cut in the grid graph.  The cut defines a
segmentation of the image into two regions, with a boundary $\Gamma$
between them.  The cut value, $\cut(A,B|G_{\text{grid}})$, counts the
number of neighboring pixels that are in different regions.  In
general the cut value in the grid graph and similar graphs can be seen
as a measure of the length of the boundary $\Gamma$ (see
\cite{boykov2003computing}).
\begin{observation}
\[\cut(A,B|G_{\text{\textnormal{grid}}}) \approx \length(\Gamma).\]
\end{observation}
This is a commonly used measure of spatial coherence in image
segmentation problems (see, e.g., \cite{mumford}).  Although the
criteria $\cut(A,B|G_{\text{grid}})$ leads to spatially coherent
segmentations and is widely used in practice, it gives most preference
to trivial solutions with a small (single pixel) region.

Using the previous observation and noting that
$\vol(S|G_{\text{grid}}) \approx 4|S|$ for $S \subseteq V$ we can
derive an expression for the value of a normalized cut in the grid
graph.
\begin{observation}
\[\ncut(A,B|G_{\textnormal{{grid}}}) \approx \frac{|V|}{4}\frac{\length(\Gamma)}{|A||B|}.\]
\end{observation}
Minimizing this criteria encourages solutions where the boundary
$\Gamma$ between the two regions is short (to minimize
$\length(\Gamma)$) and where the two regions have similar size (to
maximize $|A| |B|$).

\subsection{Global Appearance Information: $G_{\text{data}}$}

Now we consider the weight of cuts and normalized cuts in
$G_{\text{data}}$.

For $S \subseteq V$ we use $g_S$ to denote a kernel density estimate
defined by the values of pixels in $S$,
\begin{equation*}
g_S(c) = \frac{1}{|S|}\sum_{i \in S} K(I(i)-c).
\end{equation*}

\begin{proposition} 
\[\cut(A,B|G_{\text{\textnormal{data}}}) = (2\pi\sigma^2)^\frac{d}{2} |A| |B| \langle g_A, g_B\rangle.\]
\end{proposition}
Here $d$ is the dimension of $I(j)$ ($d=1$ for graylevel images and
$d=3$ for RGB images), $g_A$ and $g_B$ are densitiy estimates defined
using a Gaussian kernel, and $\langle \cdot, \cdot \rangle$ denotes
the standard inner product of functions.
\begin{proof}
  We use the fact that the convolution of two Gaussians with equal variance is a Gaussian with twice the variance,
  \begin{align*}
    \sum_{i \in A, j \in B} w_{\text{data}}(i, j) &= \sum_{i \in A}\sum_{j \in B}\exp\left( -\frac{\|| I(i) - I(j) ||^2}{2\sigma^2}\right)
    \nonumber\\
    &  = (2\pi\sigma^2)^\frac{d}{2} \sum_{i \in A}\sum_{j \in B} \int_{-\infty}^\infty
    \left(\frac{1}{\pi\sigma^2}\right)^d \exp\left(-\frac{||I(i) - c||^2}{\sigma^2}\right) \exp\left(-\frac{||I(j) - c||^2}{\sigma^2}\right)  \mathrm{d}c
    \nonumber\\
    & = (2\pi\sigma^2)^\frac{d}{2} \int_{-\infty}^\infty (\sum_{i \in A}K(I(i) - c))(\sum_{j \in B} K(I(j) - c)) \mathrm{d}c
    \nonumber\\
    &= (2\pi\sigma^2)^\frac{d}{2} |A| |B| \int_{-\infty}^\infty g_A(c) g_B(c) \mathrm{d}c = (2\pi\sigma^2)^\frac{d}{2} |A| |B| \langle g_{A}, g_{B} \rangle.
  \end{align*}
\end{proof}

The proposition above is related to the Laplacian PDF Distance in
\cite{jenssen2005laplacian}.  It is also related to the work in
\cite{tang2013grabcut} where a different graph construction was used to
define global appearance models.

The weight of a cut in $G_{\text{data}}$ will be minimized when the
pixel values in the two regions have complementary support.  Although
this intuitively makes sense, the measure encourages regions to be
unbalanced in size due to the term $|A||B|$ multiplying $\langle g_A,
g_B \rangle$.

In order to derive an expression for $\ncut(A,B|G_{\text{data}})$, we
first use a similar reasoning as in the proposition above to note that
$\vol(S|G_{\text{\textnormal{data}}}) = (2\pi\sigma^2)^{(d/2)} |S| |V|
\langle g_{S}, g_{V} \rangle$.  Then, from the definition of the
normalized cut we obtain the following result.

\begin{proposition} 
\[\ncut(A,B|G_{\text{\textnormal{data}}}) = \langle g_V, g_V \rangle \frac{\langle g_A, g_B\rangle}{\langle g_A, g_V \rangle \langle g_B, g_V \rangle}.\]
\end{proposition}
This criteria is minimized when the distributions $g_A$ and $g_B$ have
little overlap and both have significant overlap with $g_V$.  In
particular it penalizes solutions where one region does not represent
a significant amount of the image data.

\subsection{Combining Spatial and Appearance information}



The normalized cut values in $G_{\text{grid}}$ and $G_{\text{data}}$
provide complementary measures for image segmentation.  To combine the
spatial and appearance cues we use a convex combination,
\begin{equation*}
    \mixncut(A, B) = (1 - \lambda) \ncut(A, B| G_{\text{data}}) + \lambda \ncut(A, B| G_{\text{grid}}).
\end{equation*}
The parameter $\lambda \in [0,1]$ controls the relative importance of
the two normalized cut measures.

We interpret $\mixncut(A,B)$ as a mixture of an appearance and a
spatial term,
\begin{equation*}
    \mixncut(A, B) \approx (1- \lambda) \left( \langle g_V, g_V \rangle \frac{\langle g_A, g_B\rangle}{\langle g_A, g_V \rangle \langle g_B, g_V \rangle}\right) + \lambda \left(\frac{|V|}{4} \frac{\length(\Gamma)}{|A||B|}\right).
\end{equation*}
The first term encourages a partition of the image into regions with
dissimilar color distributions, while the second term encourages a
spatially coherent partition.  Both terms are normalized and avoid
biases towards solutions with small regions.  Note that each term is
normalized in a particular way that is natural and has appropriate
dimensions for the individual measures.

\section{Segmentation Algorithm}

Let $G_1$ and $G_2$ be two weighted graphs over the same set of vertices.  Now we describe a
spectral method we have used as a heuristic 
for optimizing a convex combination of two normalized cut values,
\begin{equation*}
\mixncut(A,B|G_1,G_2) = (1-\lambda)\ncut(A,B|G_1) + \lambda\ncut(A,B|G_2).
\end{equation*}

Our approach is based on the Markov chain and conductance
interpretation of normalized cuts
described in \cite{meila2001learning}.  Let $W_1$ and $W_2$ be
the weighted adjacency matrices of the two graphs while $D_1$ and
$D_2$ are the diagonal degree matrices.  Let,
\begin{align*}
P_1 & = D_1^{-1} W_1, \\
P_2 & = D_2^{-1} W_2
\end{align*}
\begin{equation}
P = (1-\lambda)P_1 + \lambda P_2.
\label{eqn:P}
\end{equation}

The matrices $P_1$ and $P_2$ define two Markov chains on $V$.  The
matrix $P$ also defines a Markov chain on $V$ where in one step we
follow $P_1$ with probability $(1-\lambda)$ and $P_2$ with probability
$\lambda$.  We compute the second largest eigenvector of $P$ to find a
cut $(A,B)$ with small conductance.

In our experiments, we use a Lanczos Process to compute the second
largest eigenvector of $P$.  We use $K$-means with $K=2$ to cluster
the entries in the eigenvector into 2 clusters.

\subsection{Graph Sparsification}
\label{sec:graph_sparse}

When the matrix $P$ is sparse we can compute the required eigenvector
much more quickly.  The grid $G_\text{grid}$ is sparse but
$G_\text{data}$ is dense.  We sparsify the graph using a random
sampling approach.

The approach described here is complementary to other methods that
have been used to speed up the computation of eigenvectors for
clustering.  One such method is based on Nystrom approximation
\cite{fowlkes2004spectral}.  Another approach involves power iteration
\cite{lin2010power}.

Let $G$ be a weighted graph.  To construct a sparse graph $G'$ we
independently sample $m$ edges (with replacement) from $G$, with
probabilities proportional to the edge weights.  The weight of each
sampled edge is set to $1$ (adding up weights if there is repetition).
With this approach the expected value of a cut $(A,B)$ in $G'$ equals
the value of the cut in $G$ up to a scaling factor of $(m/\vol(V|G))$.
Moreover, if $m$ is sufficiently large then with high probability
every cut in $G'$ has weight close to the cut value in $G$ (up to a
scaling factor of $(m/\vol(V|G))$) (see, e.g.,
\cite{williamson2011}). For the experiments in this paper we use $m =
\alpha|V|$ with $\alpha>0$ to obtain a sparse graph approximating $G$.

To implement this approach efficiently for $G_\text{data}$ we need to
sample edges with probability proportional to their weights $w(i,j)$
\emph{without} enumerating all possible edges.  We use an importance
sampling method as a practical alternative.

First we partition $V$ into $L$ ($\approx 1000$ in practice) sets
$S_1,\ldots,S_L$ with low appearance variance.  We do this greedily,
starting with a single set and repeatedly partitioning the set with
highest variance into two using the $K$-means algorithm.  Let $m_i$ be
the mean appearance of pixels in $S_i$ and
\begin{equation*}
q(a,b) = |S_a||S_b|\exp\left(-\frac{||m_a - m_b||^2}{2\sigma^2}\right).
\end{equation*}
To sample an edge for $G'$ we select a random pair $S_a$ and $S_b$
with probability proportional to $q(a,b)$.  We then select $i \in S_a$
and $j \in S_b$ uniformly at random.  Finally, we add the edge $\{i,j\}$
to $G'$ with weight $w'(i,j) = |S_a||S_b|w(i,j)/q(a,b)$.

\section{Numerical Experiments}

The algorithms in this section were implemented in MATLAB and run on a computer with
an Intel i5-6200U CPU @ 2.30GHz using 8 GB of RAM running Linux.

\subsection{Segmentation accuracy measure}

To measure the accuracy of a segmentation we use the Jaccard Index to
compare pairs of regions \cite{jaccard1901etude},
$$J(S,Q) = |S \cap Q|/|S \cup Q|.$$

Let $I$ be an image with a ground-truth binary segmentation $(S,Q)$.
Let $(A,B)$ be a the result of a binary segmentation algorithm.  To
evaluate the accuracy of $(A,B)$ we consider (1) pairing $S$ with $A$
and $Q$ with $B$ or (2) pairing $S$ with $B$ and $Q$ with $A$.  We
define the accuracy of the segmentation $(A,B)$ as,
\begin{equation*}
\jac = \max\left(\frac{J(S,A)+J(Q,B)}{2},\frac{J(S,B)+J(Q,A)}{2}\right).
\end{equation*}

\subsection{Sparsification algorithm for $\ncut$}
\begin{figure}
\centering
\begin{subfigure}[t]{0.165\linewidth}
    \centering
    \raisebox{17pt}{\includegraphics[ width=\linewidth]{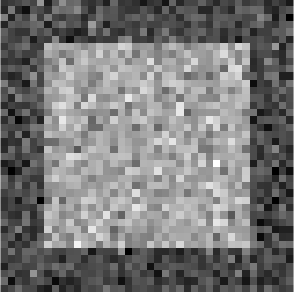}}
\end{subfigure}%
\hspace{15pt}
\begin{subfigure}[t]{0.36\linewidth}
    \centering
    \includegraphics[width=\linewidth, height=3.5cm]{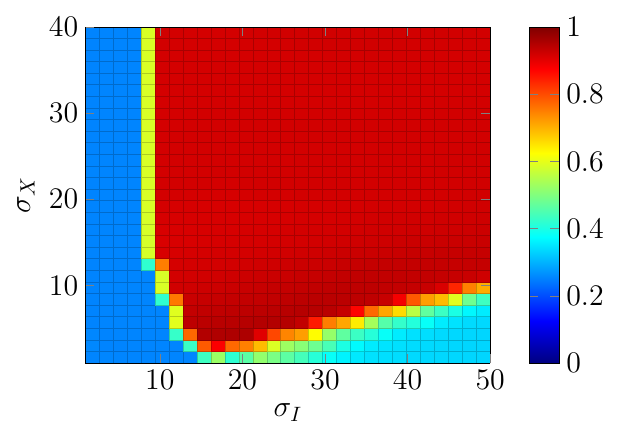}
\end{subfigure}%
\begin{subfigure}[t]{0.36\linewidth}
    \centering
    \includegraphics[width=\linewidth, height=3.5cm]{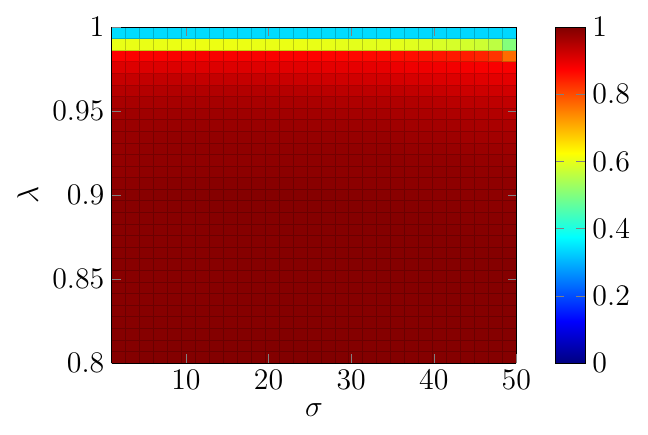}
\end{subfigure}

\begin{subfigure}[t]{0.165\linewidth}
    \centering
    \raisebox{17pt}{\includegraphics[ width=\linewidth]{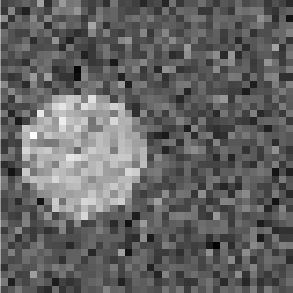}}
\end{subfigure}%
\hspace{15pt}
\begin{subfigure}[t]{0.36\linewidth}
    \centering
    \includegraphics[width=\linewidth, height=3.5cm]{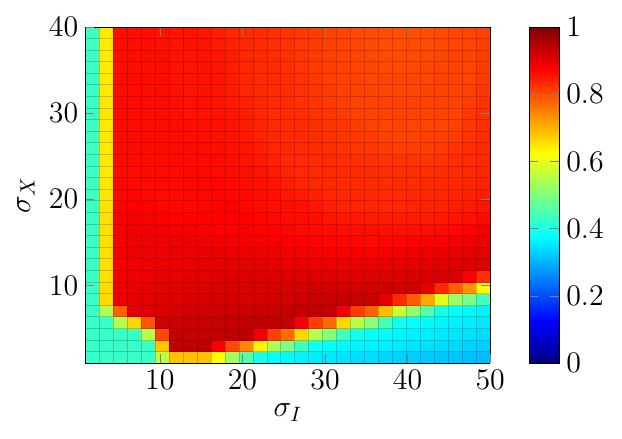}
\end{subfigure}%
\begin{subfigure}[t]{0.36\linewidth}
    \centering
    \includegraphics[width=\linewidth, height=3.5cm]{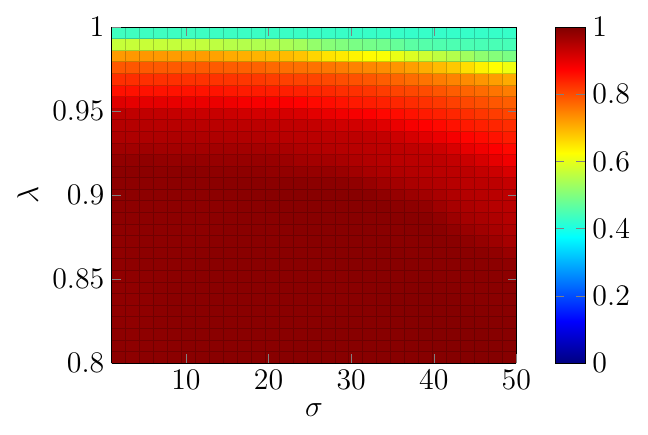}
\end{subfigure}

\begin{subfigure}[t]{0.165\linewidth}
    \centering
    \raisebox{17pt}{\includegraphics[ width=\linewidth]{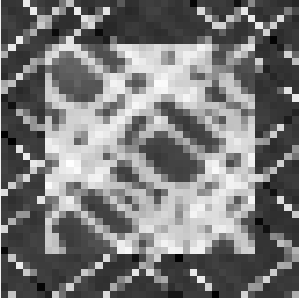}}
\end{subfigure}%
\hspace{15pt}
\begin{subfigure}[t]{0.36\linewidth}
    \centering
    \includegraphics[width=\linewidth, height=3.5cm]{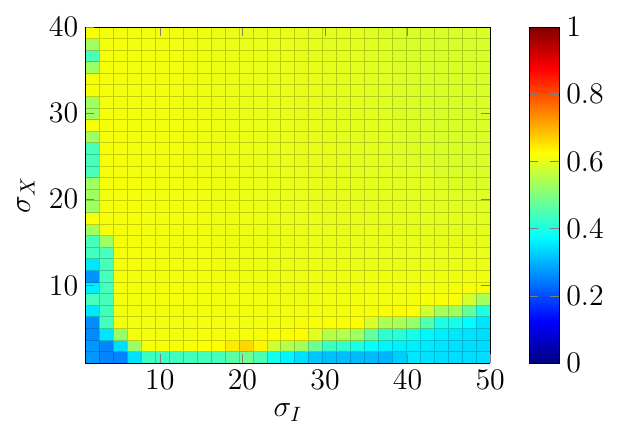}
\end{subfigure}%
\begin{subfigure}[t]{0.36\linewidth}
    \centering
    \includegraphics[width=\linewidth, height=3.5cm]{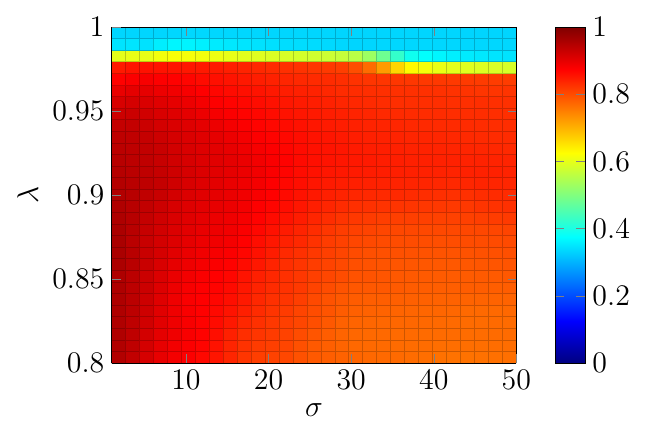}
\end{subfigure}

\begin{subfigure}[t]{0.165\linewidth}
    \centering
    \raisebox{17pt}{\includegraphics[ width=\linewidth]{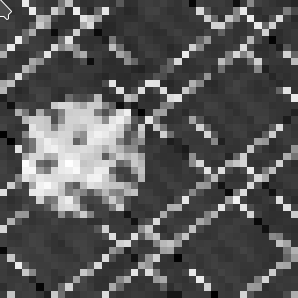}}
    \caption{Image}
\end{subfigure}%
\hspace{15pt}
\begin{subfigure}[t]{0.36\linewidth}
    \centering
    \includegraphics[width=\linewidth, height=3.5cm]{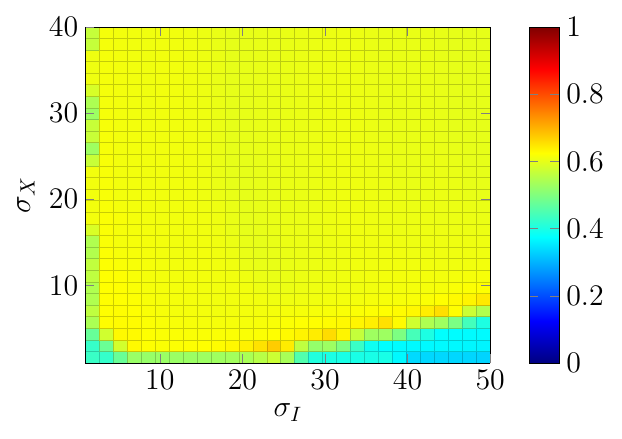}
    \caption{$\ncut$-Graylevel}
\end{subfigure}%
\begin{subfigure}[t]{0.36\linewidth}
    \centering
    \includegraphics[width=\linewidth, height=3.5cm]{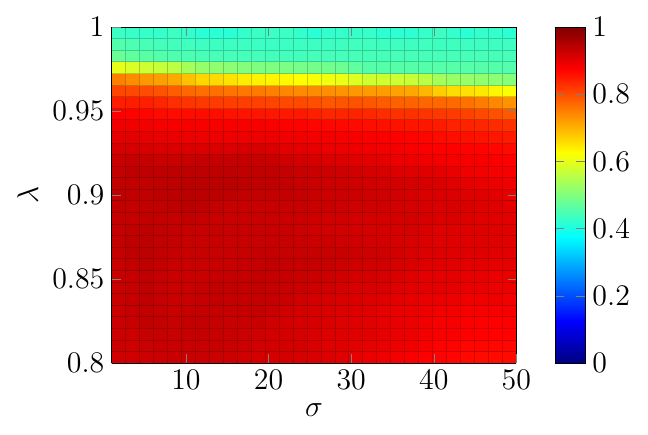}
    \caption{$\mixncut$}
\end{subfigure}
\caption{Evaluation of $\ncut$ and $\mixncut$ without graph
  sparsification on $40 \times 40$ images. Column (a) shows the input
  images.  Column (b) shows the $\jac$ value obtained with $\ncut$ for
  a range of values for $\sigma_I$ and $\sigma_X$. Column (c) shows
  the $\jac$ value obtained with $\mixncut$ over various combinations
  of $\sigma$ and $\lambda$. In these experiments $\ncut$ averaged
  $6.31 \pm 5.28$s of processing time and $\mixcut$, $1.35 \pm
  0.25$s.}\label{fig:fully_connected}
\end{figure}

We compare our new segmentation method with the original normalized
cut formulations, $\ncut$, using the graph $H$ described in
section~\ref{sec:original}.  In practice we sparsify the graph $H$ to
scale the eigenvector computation to large images. Again, we
accomplish this using importance sampling.  Let $H$ be the graph with
weights defined by Equation~(\ref{eq:trad_ncut_seg_eq}).  To sample
one edge from $H$ we first select a pixel $i$ uniformly at random.  We
then draw a location $x$ from a Normal distribution centered at $X(i)$
with variance $\sigma_{X}^2$ and select the pixel $j$ closest to that
location.  We add the edge $\{i,j\}$ to $G'$ with weight $w'(i,j) =
\exp\left(|| I(i) - I(j)||^2/2\sigma_{I}^2\right)$.  We repeat this
process $m$ times.  In the following experiments we used $m = 100 |V|$
to sparsify $H$.

\subsection{Evaluation of $\ncut$ and $\mixncut$ without sparsification}

In order to demonstrate the efficacy of our approach compared to the
traditional normalized cuts method using the graph construction given
in Equation (\ref{eq:trad_ncut_seg_eq}), we ran both $\ncut$ and
$\mixncut$ without the graph sparsification step of each
algorithm. Figure \ref{fig:fully_connected} shows the segmentation
accuracy of each method on small synthetic images under various
combinations of parameters. The input images were selected to have
regions of different sizes and textures.

These results demonstrate the effect of low values for either
$\sigma_I$ or $\sigma_X$ for the normalized cut method. In this
setting, the graph $H$ is very sparse and potentially disconnected,
which hinders the algorithm's segmentation performance.  We also see
the performance of $\ncut$ decays in textured images.

The $\mixncut$ method achieves good results in all scenarios, including in images with textured and/or unbalanced images. 

\subsection{Impact of edge sampling on $\mixncut$}
\begin{figure}
\centering
\begin{subfigure}[t]{0.14\linewidth}
    \centering
    \raisebox{7pt}{
    \includegraphics[width=\linewidth]{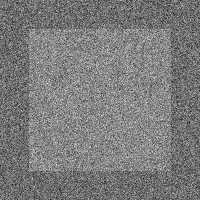}}
\end{subfigure}%
\hspace{30pt}
\begin{subfigure}[t]{0.32\textwidth}
    \centering
    \begin{tikzpicture}[trim axis left]
        \begin{axis}[
            scale only axis,
            height=2.25cm,
            width=\textwidth,
            max space between ticks=40,
            legend pos=north west,
            scaled ticks=false, 
            tick label style={/pgf/number format/fixed},
            ymax = 1,
            xtick = {},
            xticklabels = {},
            ]
            \addplot+[mark=none, smooth, black, thick, error bars/.cd,y explicit,y dir=both] table [x index=0,y index=1, y error expr=\thisrowno{2} - \thisrowno{1}] {data/varying_m_mixncut.dat};
        \end{axis}
    \end{tikzpicture}
\end{subfigure}%
\hspace{25pt}
\begin{subfigure}[t]{0.32\textwidth}
    \centering
    \begin{tikzpicture}[trim axis left]
        \begin{axis}[
            scale only axis,
            height=2.25cm,
            width=\textwidth,
            max space between ticks=40,
            legend pos=north west,
            scaled ticks=false, 
            tick label style={/pgf/number format/fixed},
            xtick = {},
            xticklabels = {},
            ]
            \addplot+[mark=none, smooth, black, thick, error bars/.cd,y explicit,y dir=both] table [x index=0,y index=4, y error expr=\thisrowno{5} - \thisrowno{4}] {data/varying_m_mixncut.dat};     
          \end{axis}
    \end{tikzpicture}
\end{subfigure}

\begin{subfigure}[t]{0.14\linewidth}
    \centering
    \raisebox{25pt}{
    \includegraphics[width=\linewidth]{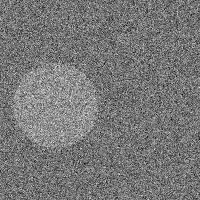}}
    \caption{Image}
\end{subfigure}%
\hspace{30pt}
\begin{subfigure}[t]{0.32\textwidth}
    \centering
    \begin{tikzpicture}[trim axis left]
        \begin{axis}[
            scale only axis,
            height=2.25cm,
            width=\textwidth,
            max space between ticks=40,
            legend pos=north west,
            scaled ticks=false, 
            tick label style={/pgf/number format/fixed},
            ymax = 1,
            xlabel = $\alpha$
            ]
            \addplot+[mark=none, smooth, black, thick, error bars/.cd,y explicit,y dir=both] table [x index=0,y index=7, y error expr=\thisrowno{8} - \thisrowno{7}] {data/varying_m_mixncut.dat};        
        \end{axis}
    \end{tikzpicture}
    \caption{Jaccard}
\end{subfigure}%
\hspace{25pt}
\begin{subfigure}[t]{0.32\textwidth}
    \centering
    \begin{tikzpicture}[trim axis left]
        \begin{axis}[
            scale only axis,
            height=2.25cm,
            width=\textwidth,
            max space between ticks=40,
            legend pos=north west,
            scaled ticks=false, 
            tick label style={/pgf/number format/fixed},
            xlabel = $\alpha$
            ]
            \addplot+[mark=none, smooth, black, thick, error bars/.cd,y explicit,y dir=both] table [x index=0,y index=10, y error expr=\thisrowno{11} - \thisrowno{10}] {data/varying_m_mixncut.dat};       
          \end{axis}
    \end{tikzpicture}
    \caption{Time (s)}
\end{subfigure}
\caption{Impact of graph sparsification using various values for
  $\alpha$. Column (a) shows two different test images of size $200
  \times 200$. Column (b) shows the average and standard deviation of
  the $\jac$ accuracy measure over 100 runs of $\mixcut$ with
  randomized graph sparsification.  Column (c) shows the average and
  standard deviation of the running time of the algorithm.  Here, we
  set $\lambda = 0.95$ and $\sigma = 1$.}\label{fig:vary_m}
\end{figure}

Figure \ref{fig:vary_m} shows how our algorithm performs under
different values of $\alpha$, where $m = \alpha |V|$ is the number of
sampled edges of $G_{\text{data}}$ when sparsified according to the
algorithm in Section \ref{sec:graph_sparse}. Our method obtains
satisfactory results in terms of $\jac$ index for $\alpha \geq 2$ and
achieves almost perfect segmentations for when $\alpha \geq
4$. Furthermore, our method has the lowest processing time when
$\alpha$ is close to 2. The increase in computation time for values of
$\alpha$ smaller that $2$ is due to the slow convergence of the
Lanczos algorithm in that regime. Having that in mind, the number of
sampled edges used to sparsify $G_{\text{data}}$ was set to $m=2|V|$
for the experiments with $\mixncut$ in the following sections.

\subsection{The role of $\lambda$ on $\mixncut$}

\begin{figure}
    \begin{subfigure}[t]{0.131\textwidth}
        \includegraphics[width=\linewidth]{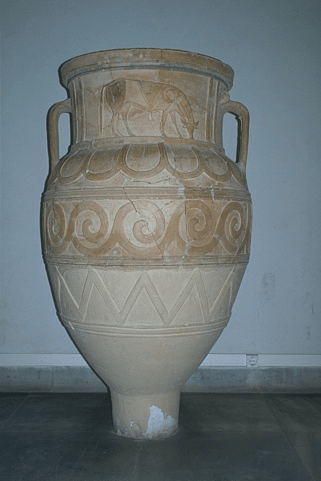}
    \end{subfigure}
    \hspace{4pt}
    \begin{subfigure}[t]{0.262\textwidth}
        \includegraphics[width=.5\linewidth]{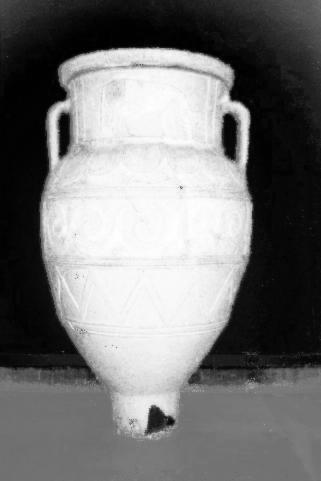}~%
        {\includegraphics[width=.5\linewidth]{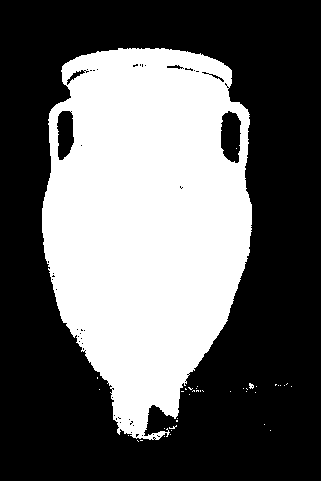}}
    \end{subfigure}%
    \hspace{8pt}
    \begin{subfigure}[t]{0.262\textwidth}
        \includegraphics[width=.5\linewidth]{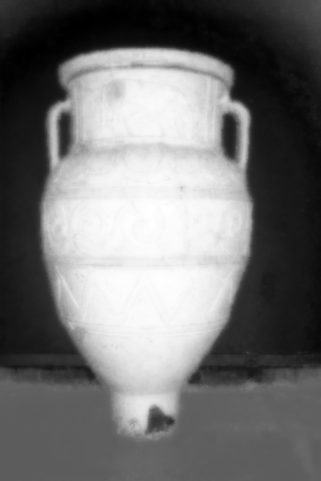}~%
        {\includegraphics[width=.5\linewidth]{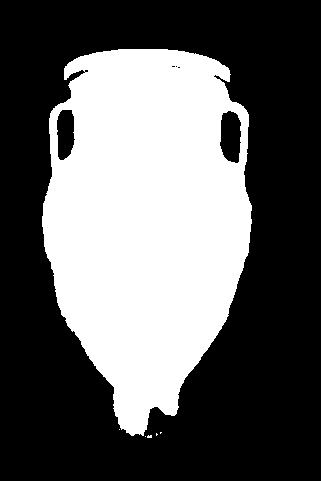}}
    \end{subfigure}%
    \hspace{8pt}
    \begin{subfigure}[t]{0.262\textwidth}
        \includegraphics[width=.5\linewidth]{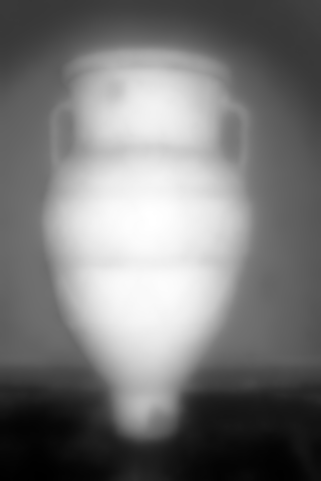}~%
        {\includegraphics[width=.5\linewidth]{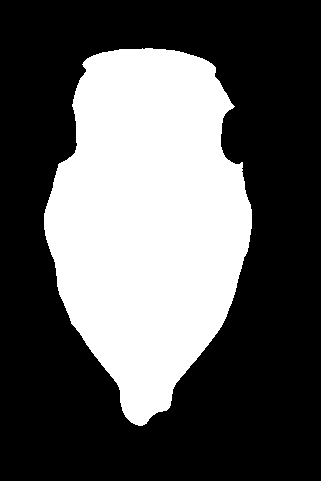}}
    \end{subfigure}

    \begin{subfigure}[t]{0.131\textwidth}
        \includegraphics[width=\linewidth]{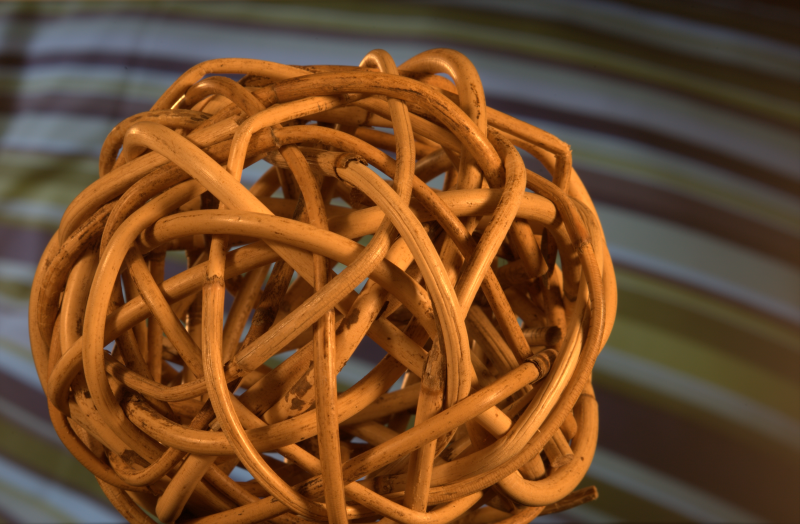}
        \caption{Image}
    \end{subfigure}
    \hspace{4pt}
    \begin{subfigure}[t]{0.262\textwidth}
        \includegraphics[width=.5\linewidth]{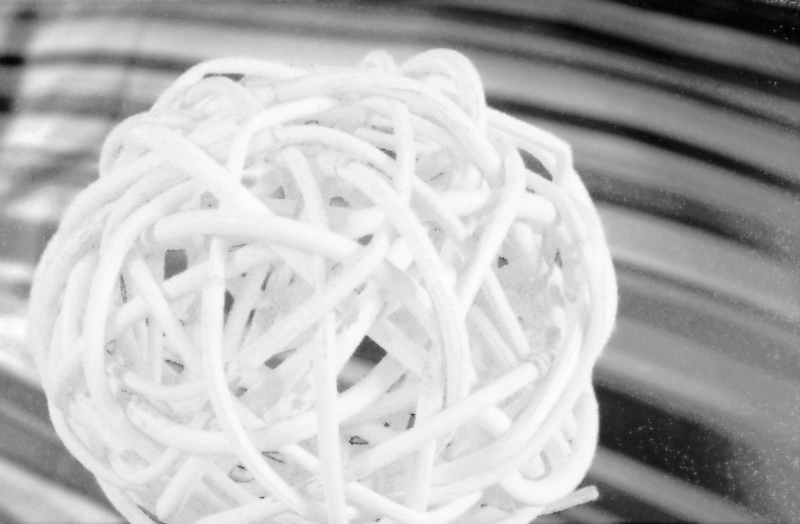}~%
        {\includegraphics[width=.5\linewidth]{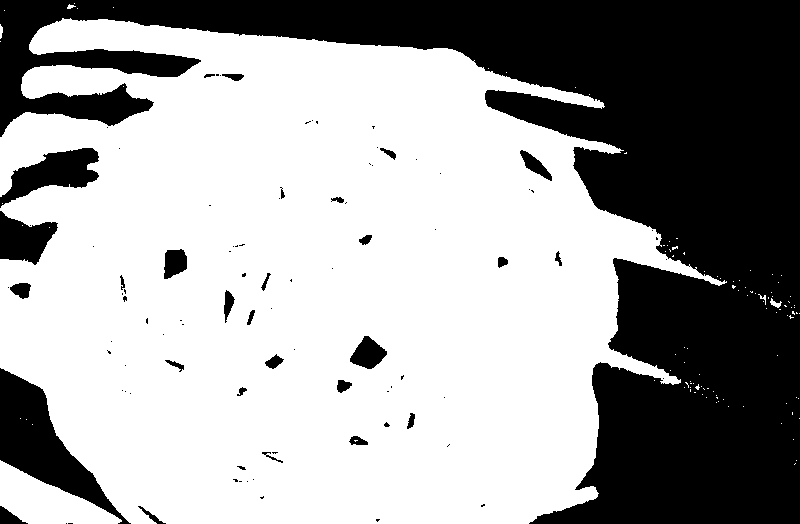}}
        \caption{$\lambda = 0.5$}
    \end{subfigure}%
    \hspace{8pt}
    \begin{subfigure}[t]{0.262\textwidth}
        \includegraphics[width=.5\linewidth]{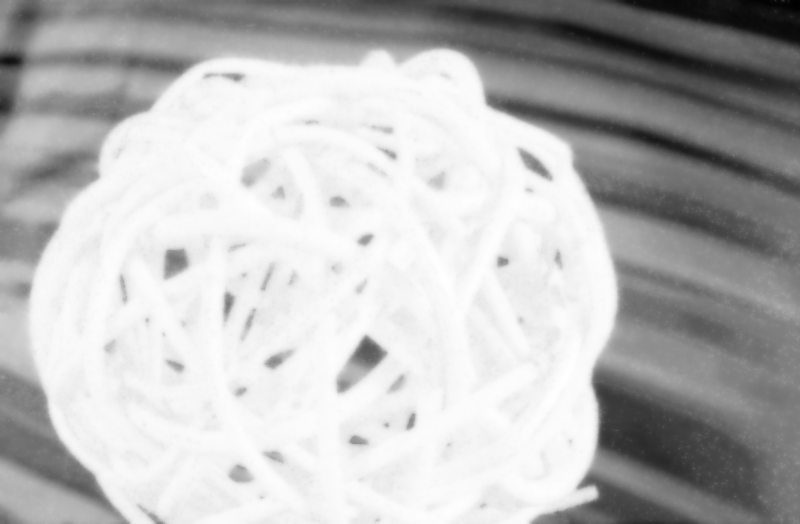}~%
        {\includegraphics[width=.5\linewidth]{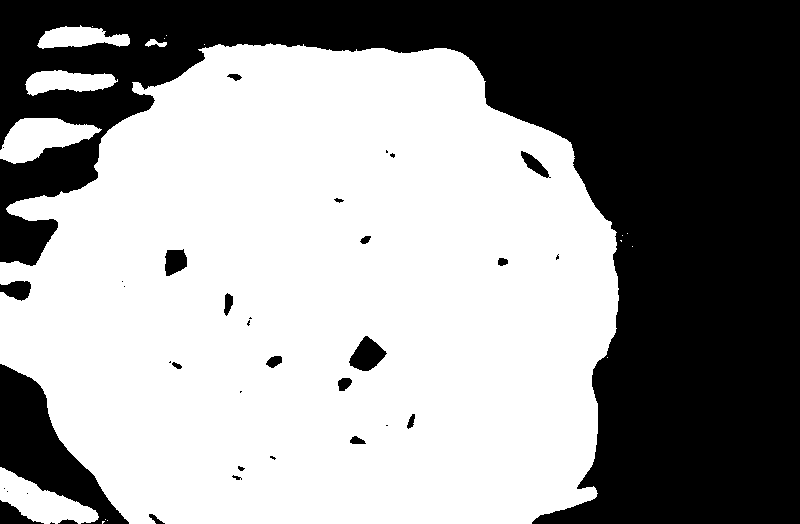}}
        \caption{$\lambda=0.9$}
    \end{subfigure}%
    \hspace{8pt}
    \begin{subfigure}[t]{0.262\textwidth}
        \includegraphics[width=.5\linewidth]{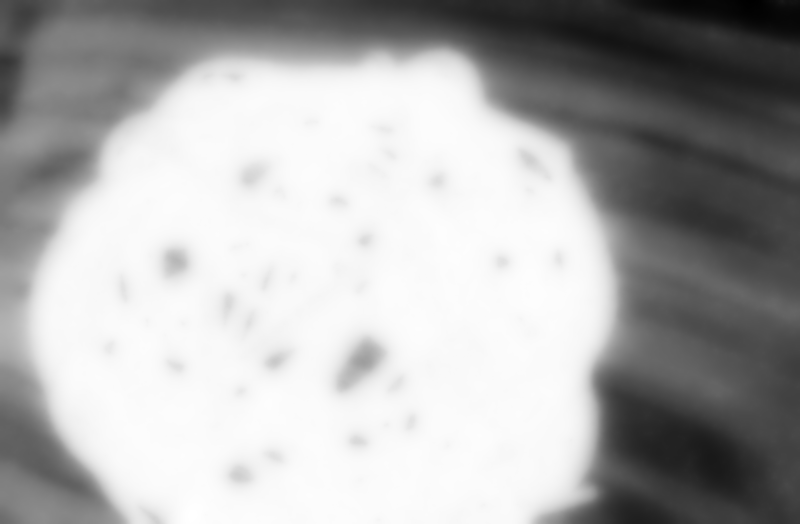}~%
        {\includegraphics[width=.5\linewidth]{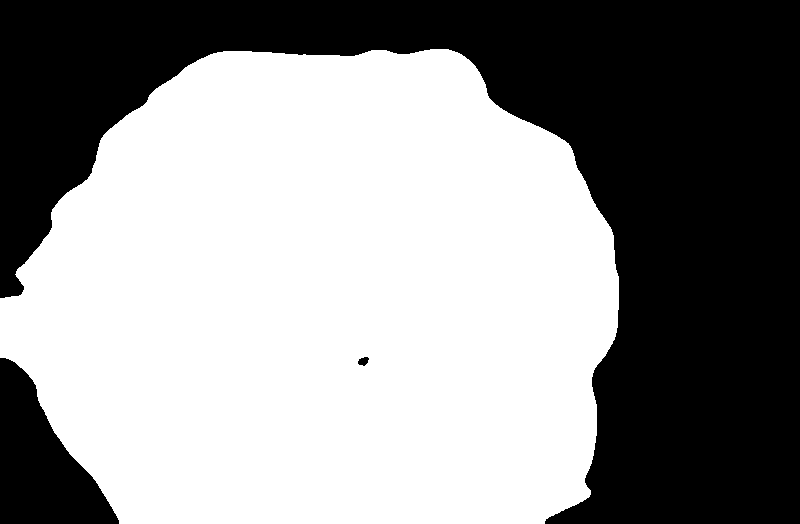}}
        \caption{$\lambda=0.99$}
    \end{subfigure}

    \caption{$\mixncut$ results. Column (a) shows the input image. Columns (b)-(d) show the eigenvectors (on the left) and segmentations (on the right) given by $\mixncut$ for various values of $\lambda$ when $\sigma = 1$.}
    \label{fig:vary_lambda}
\end{figure}

Figure \ref{fig:vary_lambda} depicts the impact of varying $\lambda$
in our method. When $\lambda$ is smaller $\mixcut$ outputs a
segmentation where fine image structures are preserved.  As $\lambda$
increases, the resulting eigenvector is blurred, leading to
segmentations with without holes or small/thin structures.  These
results demonstrate the role that $\lambda$, and therefore
$G_{\text{grid}}$, plays to enforce spatial coherence within each
region, whereas $G_{\text{data}}$ promotes the fitting of the color
data in each region.

\subsection{Experiments in Real Images}

\begin{figure}
        \begin{subfigure}[t]{0.19\textwidth}
        \includegraphics[width=.98\linewidth]{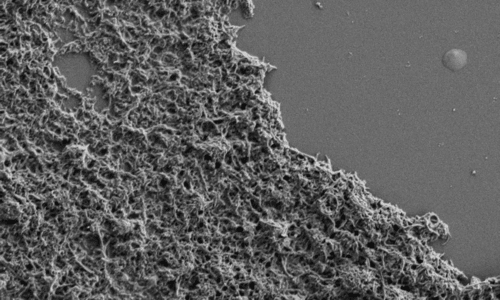}
    \end{subfigure}
    \hspace{3pt}
    \begin{subfigure}[t]{0.38\textwidth}
        \includegraphics[width=.49\linewidth]{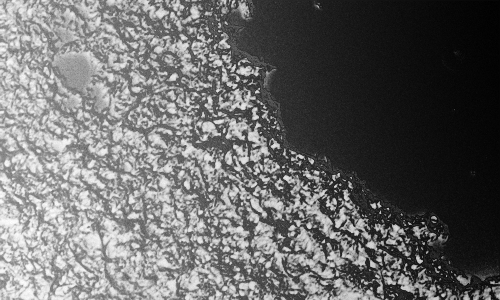}~%
        \includegraphics[width=.49\linewidth]{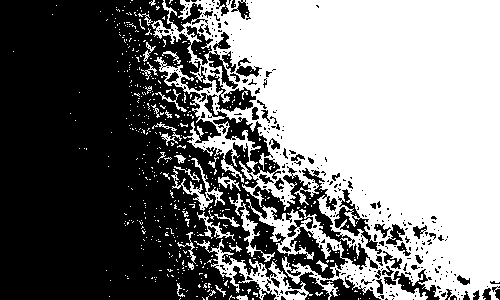}
    \end{subfigure}%
    \hspace{6pt}
    \begin{subfigure}[t]{0.38\textwidth}
        \includegraphics[width=.49\linewidth]{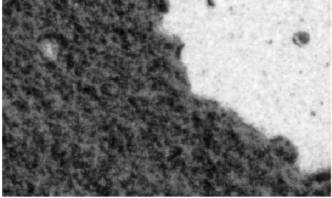}~%
        \includegraphics[width=.49\linewidth]{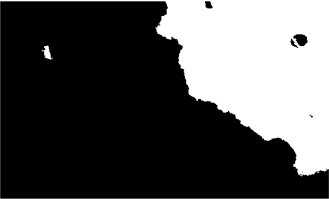}
    \end{subfigure}

    \begin{subfigure}[t]{0.19\textwidth}
        \includegraphics[width=.98\linewidth]{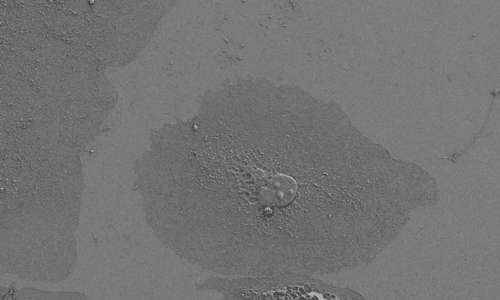}
    \end{subfigure}
    \hspace{3pt}
    \begin{subfigure}[t]{0.38\textwidth}
        \includegraphics[width=.49\linewidth]{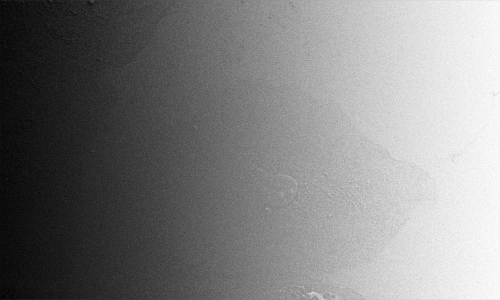}~%
        \includegraphics[width=.49\linewidth]{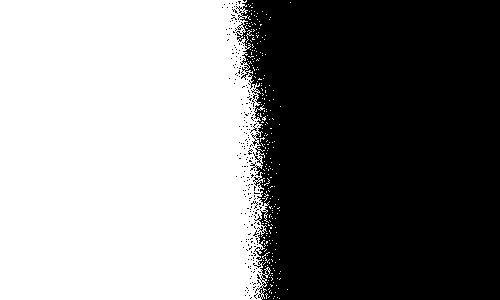}
    \end{subfigure}%
    \hspace{6pt}
    \begin{subfigure}[t]{0.38\textwidth}
        \includegraphics[width=.49\linewidth]{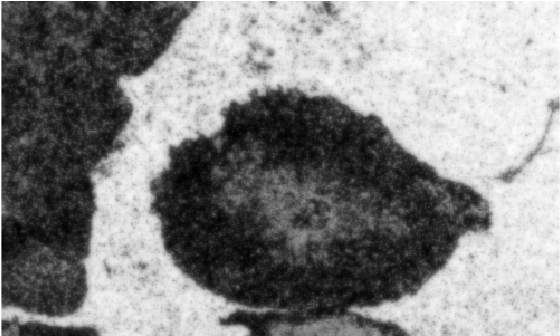}~%
        \includegraphics[width=.49\linewidth]{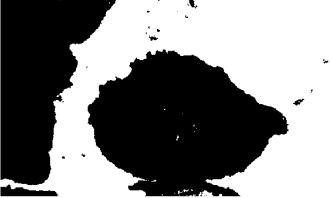}
    \end{subfigure}

    \begin{subfigure}[t]{0.19\textwidth}
        \includegraphics[width=.98\linewidth]{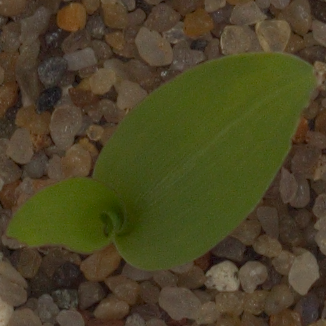}
    \end{subfigure}
    \hspace{3pt}
    \begin{subfigure}[t]{0.38\textwidth}
        \includegraphics[width=.49\linewidth]{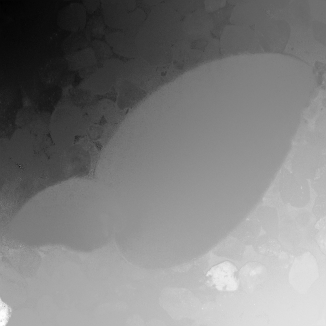}~%
        \includegraphics[width=.49\linewidth]{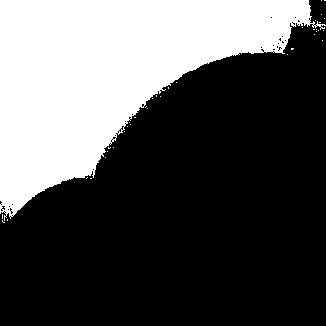}
    \end{subfigure}%
    \hspace{6pt}
    \begin{subfigure}[t]{0.38\textwidth}
        \includegraphics[width=.49\linewidth]{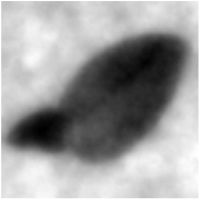}~%
        \includegraphics[width=.49\linewidth]{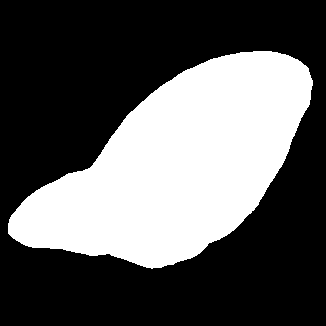}
    \end{subfigure}
        
    \begin{subfigure}[t]{0.19\textwidth}
        \includegraphics[width=.98\linewidth]{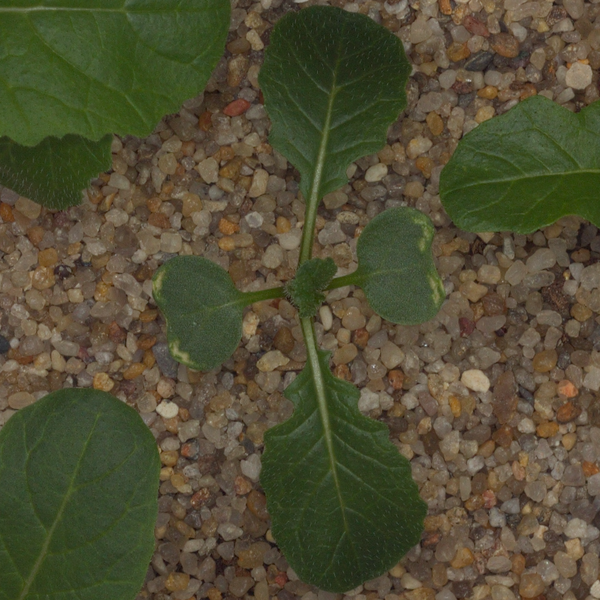}
    \end{subfigure}
    \hspace{3pt}
    \begin{subfigure}[t]{0.38\textwidth}
        \includegraphics[width=.49\linewidth]{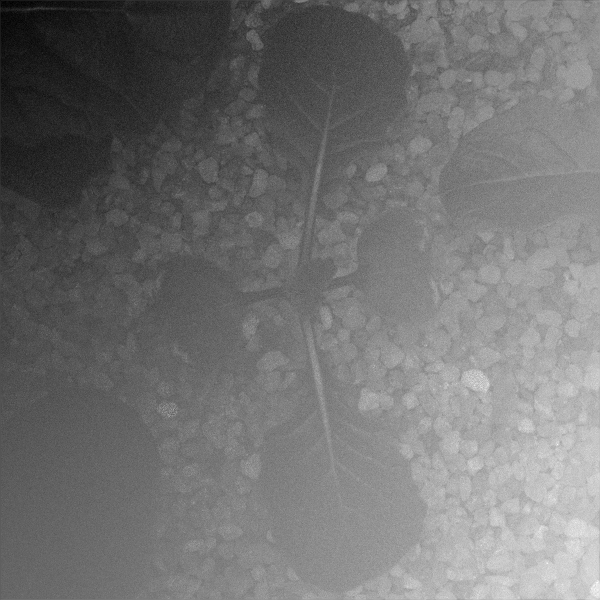}~%
        \includegraphics[width=.49\linewidth]{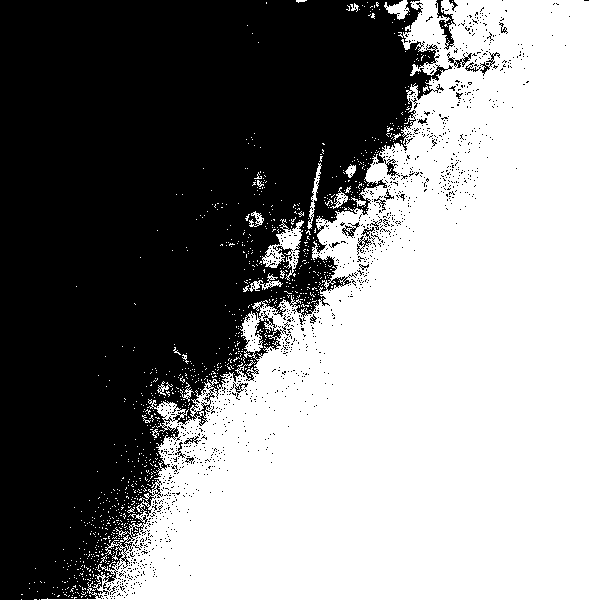}
    \end{subfigure}%
    \hspace{6pt}
    \begin{subfigure}[t]{0.38\textwidth}
        \includegraphics[width=.49\linewidth]{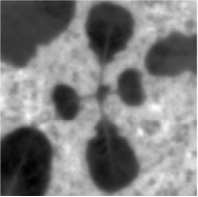}~%
        \includegraphics[width=.49\linewidth]{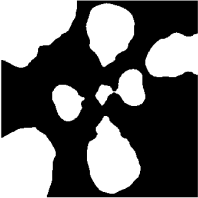}
    \end{subfigure}

    \begin{subfigure}[t]{0.19\textwidth}
        \includegraphics[width=.98\linewidth]{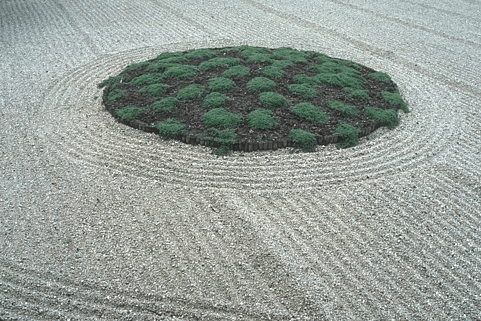}
    \end{subfigure}
    \hspace{3pt}
    \begin{subfigure}[t]{0.38\textwidth}
        \includegraphics[width=.49\linewidth]{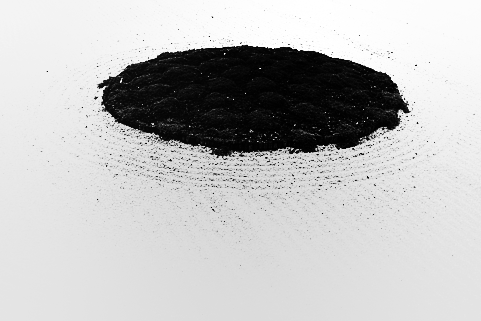}~%
        \includegraphics[width=.49\linewidth]{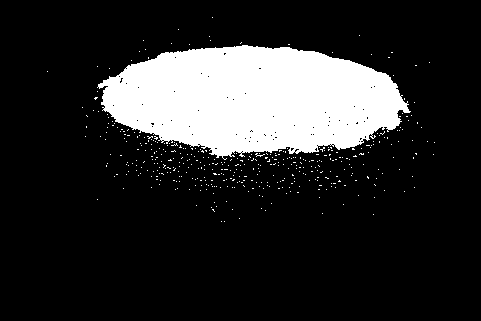}
    \end{subfigure}%
    \hspace{6pt}
    \begin{subfigure}[t]{0.38\textwidth}
        \includegraphics[width=.49\linewidth]{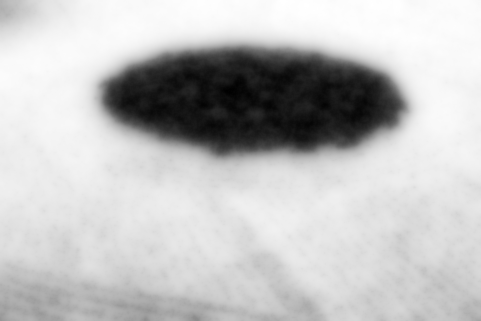}~%
        \includegraphics[width=.49\linewidth]{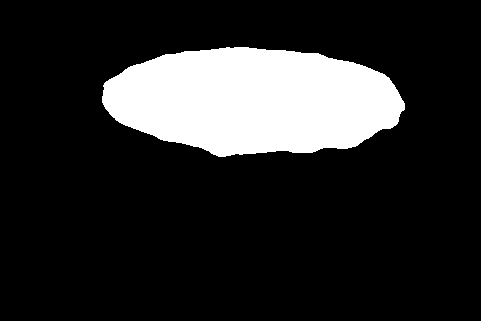}
    \end{subfigure}
    
    \begin{subfigure}[t]{0.19\textwidth}
        \includegraphics[width=.98\linewidth]{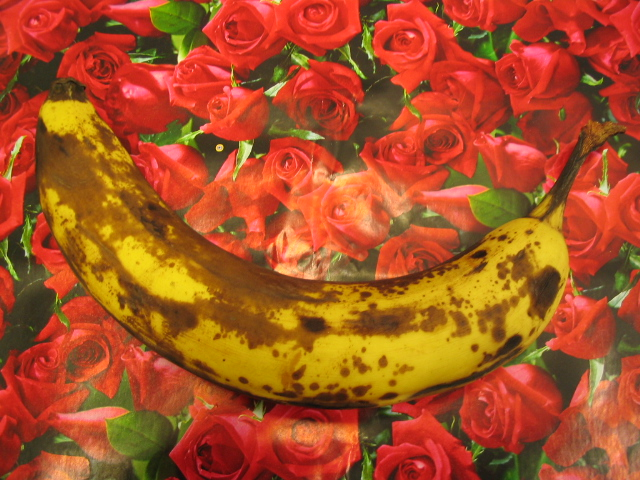}
    \end{subfigure}
    \hspace{3pt}
    \begin{subfigure}[t]{0.38\textwidth}
        \includegraphics[width=.49\linewidth]{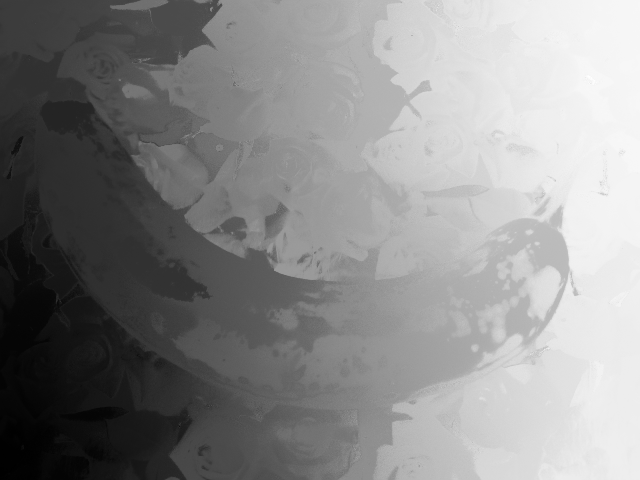}~%
        \includegraphics[width=.49\linewidth]{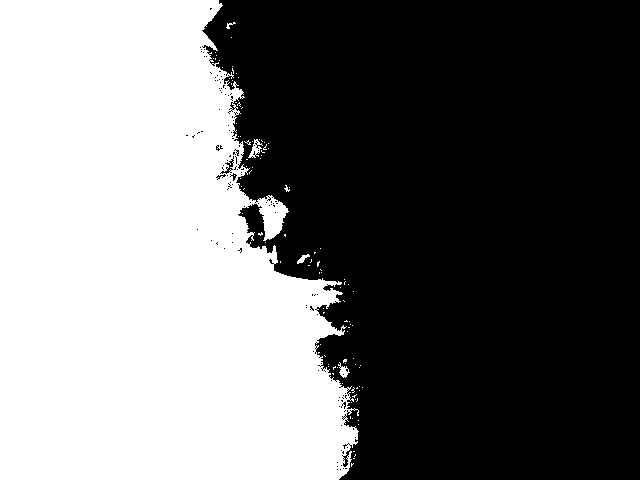}
    \end{subfigure}%
    \hspace{6pt}
    \begin{subfigure}[t]{0.38\textwidth}
        \includegraphics[width=.49\linewidth]{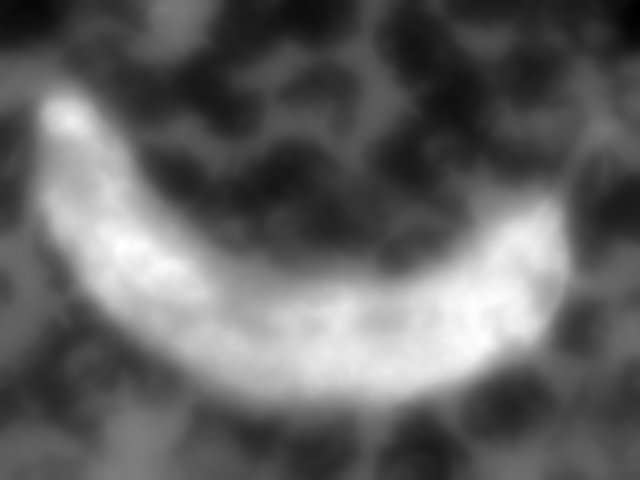}~%
        \includegraphics[width=.49\linewidth]{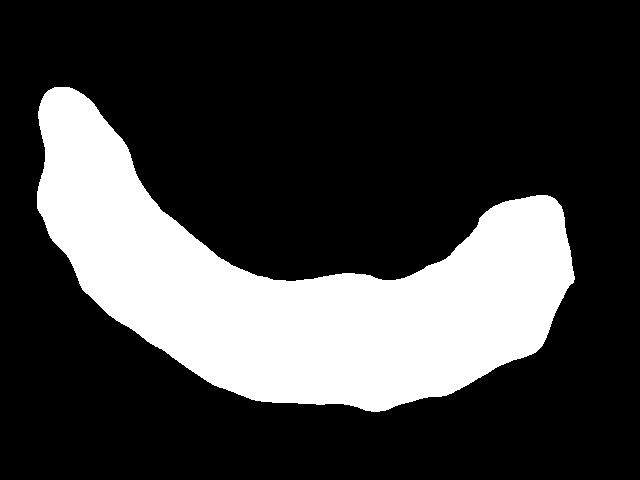}
    \end{subfigure}
        
    \begin{subfigure}[t]{0.19\textwidth}
        \includegraphics[width=.98\linewidth]{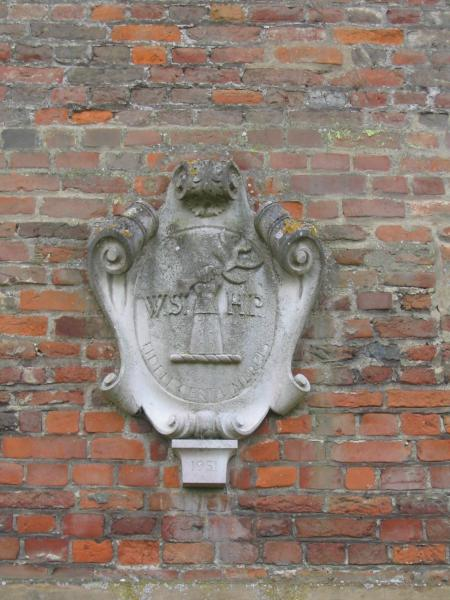}
    \end{subfigure}
    \hspace{3pt}
    \begin{subfigure}[t]{0.38\textwidth}
        \includegraphics[width=.49\linewidth]{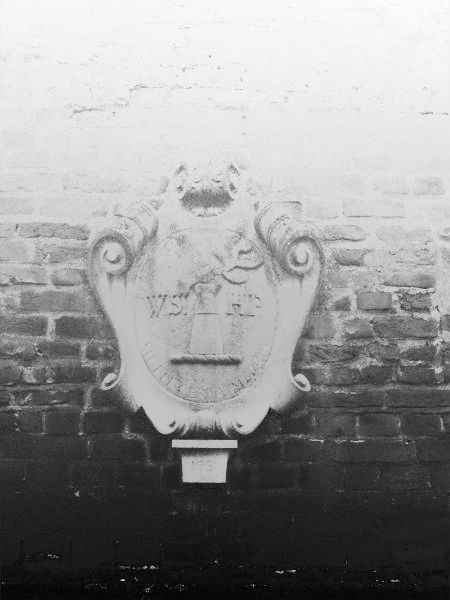}~%
        \includegraphics[width=.49\linewidth]{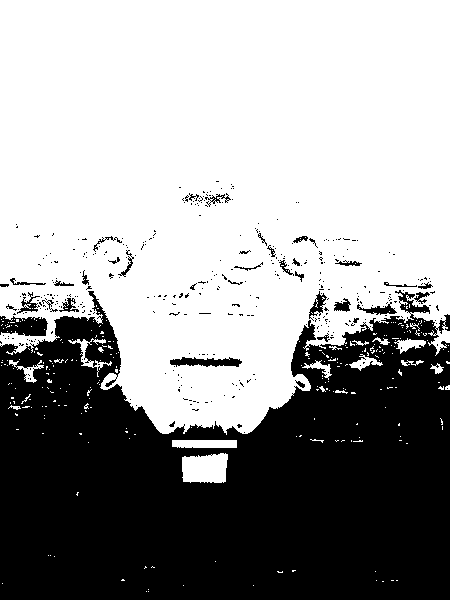}
    \end{subfigure}%
    \hspace{6pt}
    \begin{subfigure}[t]{0.38\textwidth}
        \includegraphics[width=.49\linewidth]{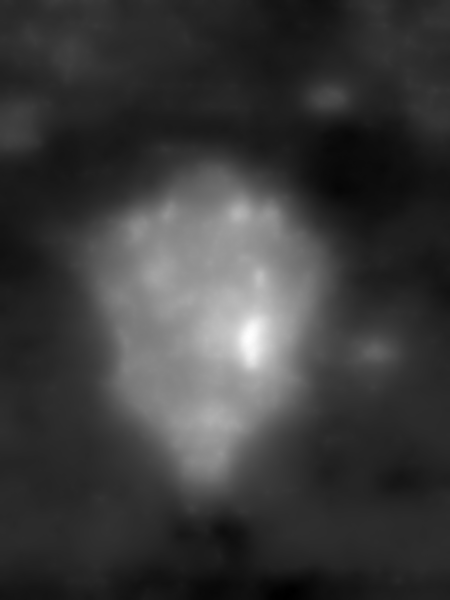}~%
        \includegraphics[width=.49\linewidth]{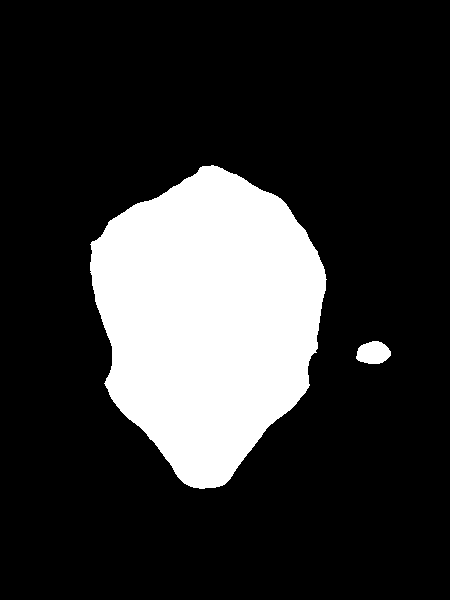}
    \end{subfigure}
        
    \begin{subfigure}[t]{0.19\textwidth}
        \includegraphics[width=.98\linewidth]{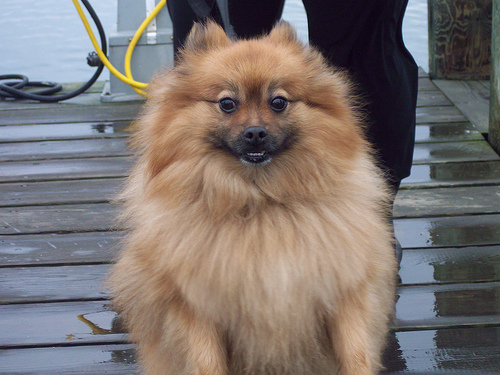}
    \end{subfigure}
    \hspace{3pt}
    \begin{subfigure}[t]{0.38\textwidth}
        \includegraphics[width=.49\linewidth]{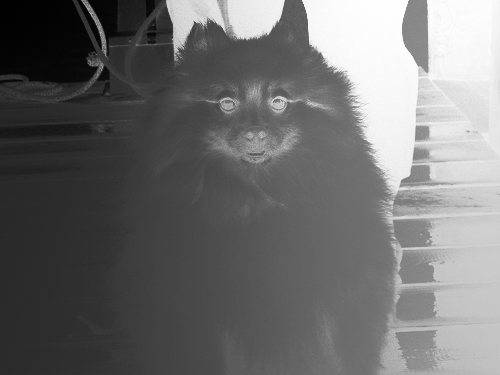}~%
        \includegraphics[width=.49\linewidth]{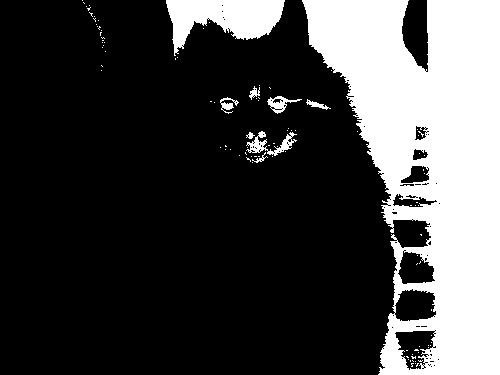}
    \end{subfigure}%
    \hspace{6pt}
    \begin{subfigure}[t]{0.38\textwidth}
        \includegraphics[width=.49\linewidth]{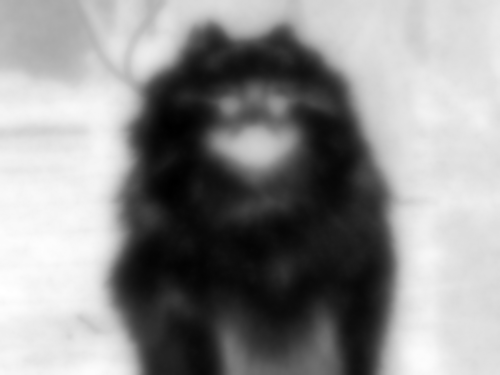}~%
        \includegraphics[width=.49\linewidth]{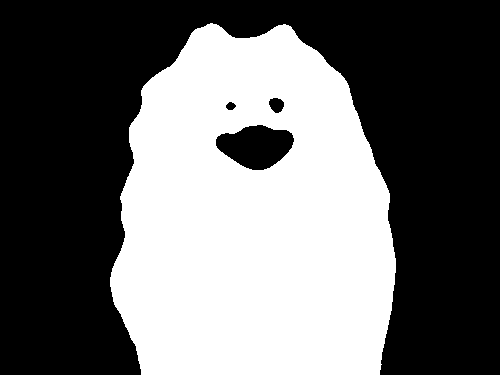}
    \end{subfigure}
    
    
    \caption{Segmentation results.  
      Column (a) shows the input images.  Column (b)
      shows the eigenvector found by the original $\ncut$ formulation
      on the left and the segmentation result on the right.  Column
      (c) shows the eigenvector and segmentation found by the $\mixncut$ method.}\label{fig:real_exp}
\end{figure}

We tested our method on real images from a variety of datasets
including the Berkeley Segmentation Dataset \cite{martin01}, the Plant
Seedlings Dataset \cite{giselsson2017public}, the Grabcut dataset
\cite{rother2004grabcut}, the PASCAL VOC dataset
\cite{everingham2015pascal} and a Scanning Electron Microscope (SEM)
dataset \cite{semdataset2018}.  Figure \ref{fig:real_exp} shows some
of the results we obtained, comparing the original normalized cuts
formulation with our new approach.  We can see in these examples how
the new approach can segment challenging images in a variety of
settings, often outperforming the original normalized cuts
formulation.

\begin{figure}
\centering
    \begin{subfigure}[t]{0.26\textwidth}
        \includegraphics[width = \textwidth, height = 2.5cm]{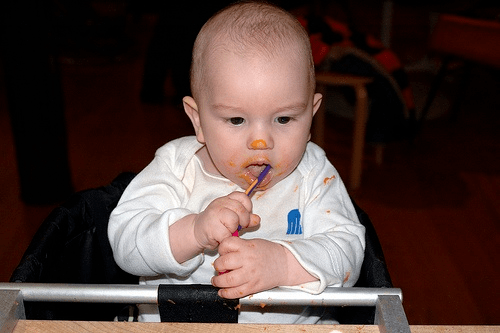}\\
        \includegraphics[width = \textwidth, height = 2.5cm]{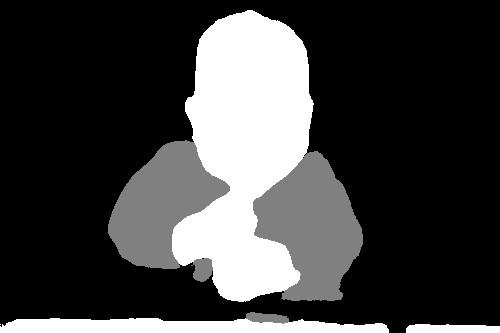}
    \end{subfigure}
    \hspace{1pt}
    \begin{subfigure}[t]{0.30\textwidth}
        \includegraphics[width = \textwidth, height = 2.5cm]{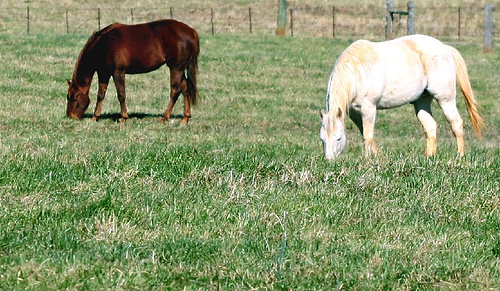}\\
        \includegraphics[width = \textwidth, height = 2.5cm]{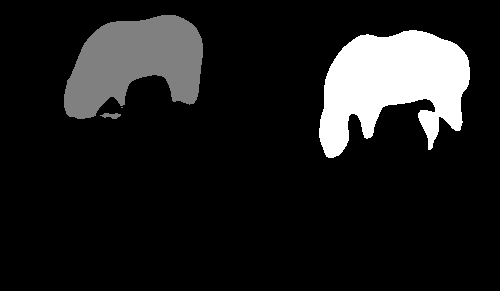}
    \end{subfigure}
    \hspace{1pt}
    \begin{subfigure}[t]{0.13\textwidth}
        \includegraphics[width = \textwidth, height = 2.5cm]{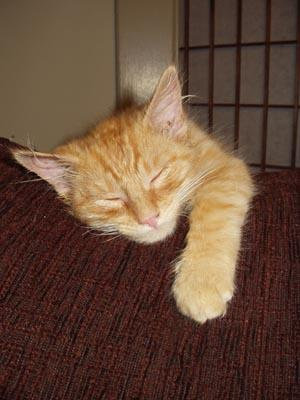}\\
        \includegraphics[width = \textwidth, height = 2.5cm]{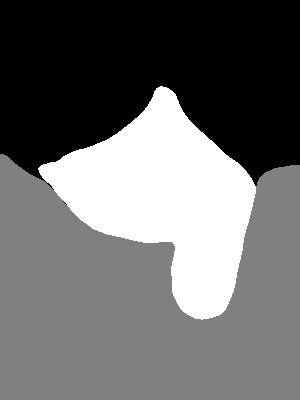}
    \end{subfigure}
    \hspace{1pt}
    \begin{subfigure}[t]{0.26\textwidth}
        \includegraphics[width = \textwidth, height = 2.5cm]{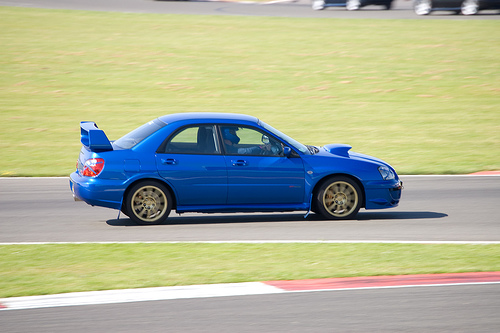}\\
        \includegraphics[width = \textwidth, height = 2.5cm]{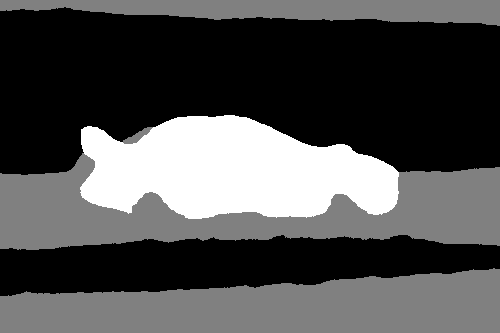}
    \end{subfigure}
    \caption{Segmentation results using the proposed method for images
      with more than 2 regions.}
    \label{fig:3_regions}
\end{figure}

Figure \ref{fig:3_regions} illustrates segmentation results using
$\mixncut$ to partition an image into 3 regions.  In this case we
follow the approach suggested in \cite{ng2002} and
\cite{meila2001learning}, using $K$-means with $K=3$ to cluster the
pixels using the second and third largest eigenvector of the
transition matrix $P$ in Equation~(\ref{eqn:P}).

For each example in these figures, we ran the algorithms using
different parameter values (specified in the next section), and show
the best result among the different runs.

\subsection{Experiments in Synthetic Images}

\begin{figure}
\begin{subfigure}[t]{0.131\textwidth}
        \includegraphics[width=\linewidth]{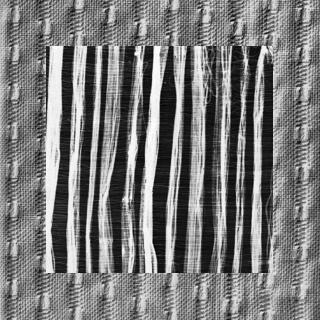}
    \end{subfigure}
    \hspace{4pt}
    \begin{subfigure}[t]{0.262\textwidth}
        \includegraphics[width=.5\linewidth]{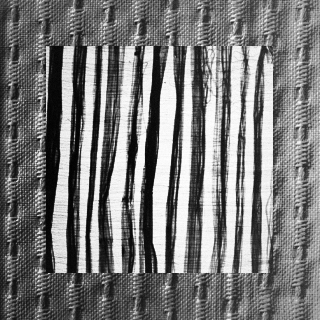}~%
        \frame{\includegraphics[width=.5\linewidth]{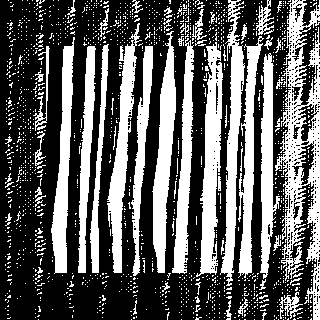}}
    \end{subfigure}%
    \hspace{8pt}
    \begin{subfigure}[t]{0.262\textwidth}
        \includegraphics[width=.5\linewidth]{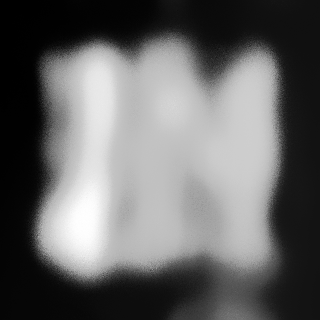}~%
        \frame{\includegraphics[width=.5\linewidth]{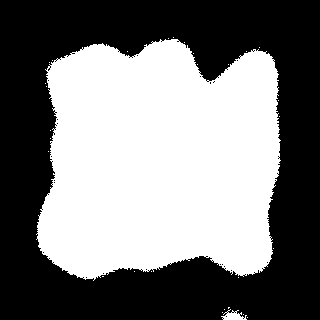}}
    \end{subfigure}%
    \hspace{8pt}
    \begin{subfigure}[t]{0.262\textwidth}
        \includegraphics[width=.5\linewidth]{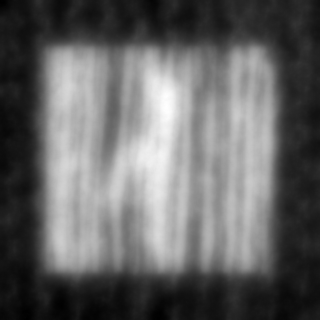}~%
        \frame{\includegraphics[width=.5\linewidth]{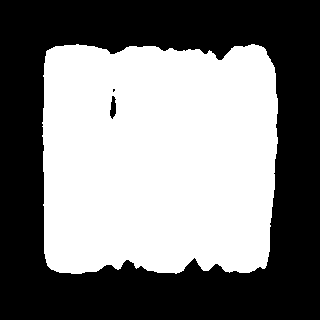}}
    \end{subfigure}
    
    \vspace{5pt}
    
    \begin{subfigure}[t]{0.131\textwidth}
        \includegraphics[width=\linewidth]{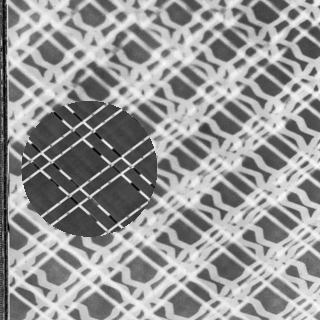}
    \end{subfigure}
    \hspace{4pt}
    \begin{subfigure}[t]{0.262\textwidth}
        \includegraphics[width=.5\linewidth]{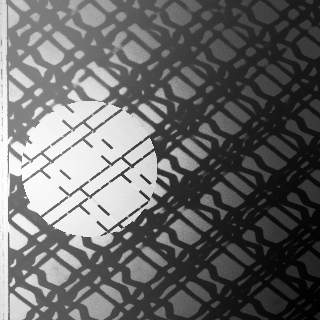}~%
        \frame{\includegraphics[width=.5\linewidth]{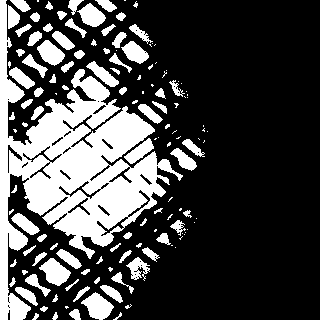}}
    \end{subfigure}%
    \hspace{8pt}
    \begin{subfigure}[t]{0.262\textwidth}
        \includegraphics[width=.5\linewidth]{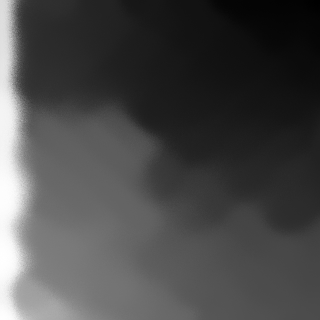}~%
        \frame{\includegraphics[width=.5\linewidth]{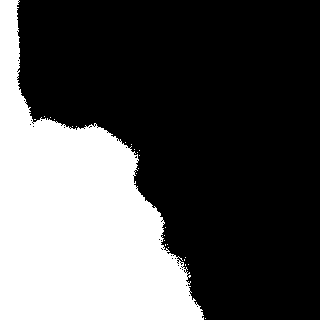}}
    \end{subfigure}%
    \hspace{8pt}
    \begin{subfigure}[t]{0.262\textwidth}
        \includegraphics[width=.5\linewidth]{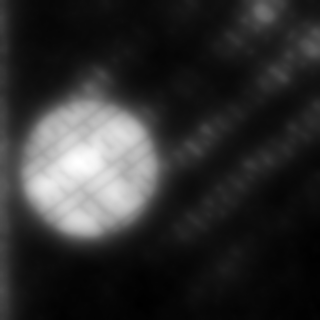}~%
        \frame{\includegraphics[width=.5\linewidth]{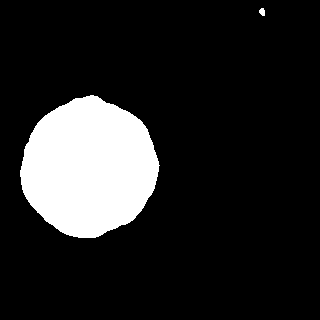}}
    \end{subfigure}
    
    \vspace{5pt}
    
    \begin{subfigure}[t]{0.131\textwidth}
        \includegraphics[width=\linewidth]{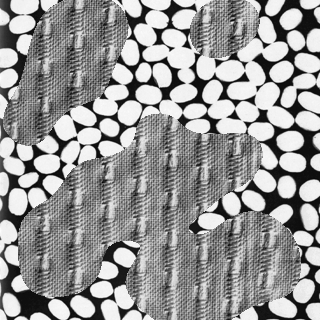}
        \caption{Image}
    \end{subfigure}
    \hspace{4pt}
    \begin{subfigure}[t]{0.262\textwidth}
        \includegraphics[width=.5\linewidth]{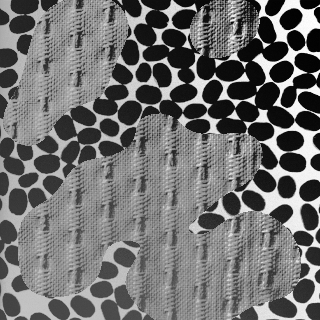}~%
        \frame{\includegraphics[width=.5\linewidth]{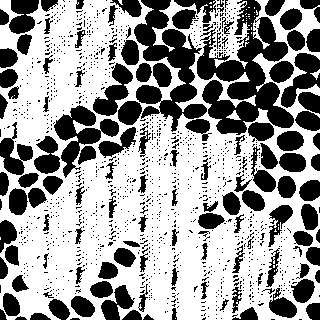}}
        \caption{$\ncut$-Graylevel}\label{fig:brodatz_examples_seg_ncut}
    \end{subfigure}%
    \hspace{8pt}
    \begin{subfigure}[t]{0.262\textwidth}
        \includegraphics[width=.5\linewidth]{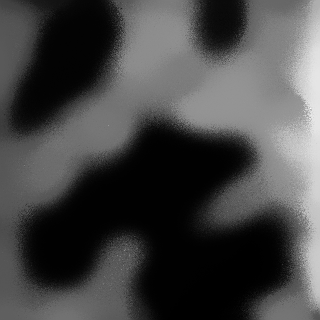}~%
        \frame{\includegraphics[width=.5\linewidth]{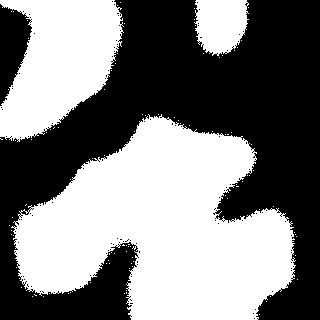}}
        \caption{$\ncut$-Gabor}\label{fig:brodatz_examples_seg_gabor}
    \end{subfigure}%
    \hspace{8pt}
    \begin{subfigure}[t]{0.262\textwidth}
        \includegraphics[width=.5\linewidth]{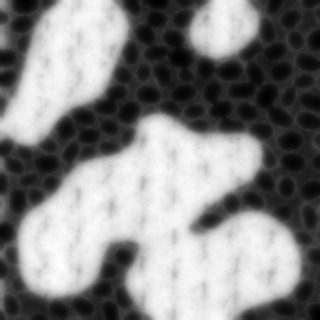}~%
        \frame{\includegraphics[width=.5\linewidth]{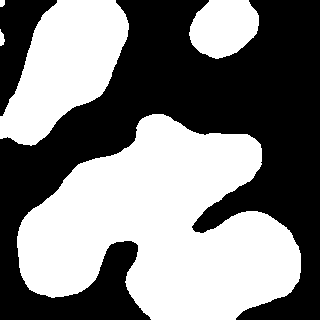}}
        \caption{$\mixcut$}\label{fig:brodatz_examples_seg_mixcut}
    \end{subfigure}

    \caption{Comparing $\ncut$-Graylevel, $\ncut$-Gabor, and $\mixncut$ on
      textured images.  Column (a) shows the input images.  Column (b)
      shows the eigenvector found by the original $\ncut$ formulation
      on the left and the segmentation result on the right.  Column
      (c) shows the eigenvector found by $\ncut$ with Gabor features
      on the left and the segmentation result on the right.  Column
      (d) shows the eigenvector found by the new $\mixncut$
      formulation on the left and the segmentation result on the
      right.}
    \label{fig:brodatz_examples_seg}
\end{figure}

For a quantitative evaluation we used images with Brodatz textures
\cite{brodatz1966textures}.  To generate input images, we mixed pairs
of textures using different ground-truth segmentation patterns.

\begin{figure}
    \begin{subfigure}[t]{0.095\textwidth}
        \includegraphics[width=\linewidth]{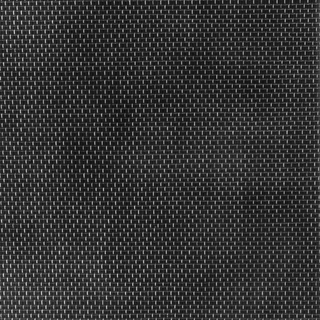}
    \end{subfigure}
    \begin{subfigure}[t]{0.095\textwidth}
        \includegraphics[width=\linewidth]{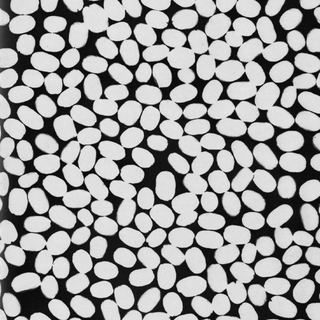}
    \end{subfigure}
    \begin{subfigure}[t]{0.095\textwidth}
        \includegraphics[width=\linewidth]{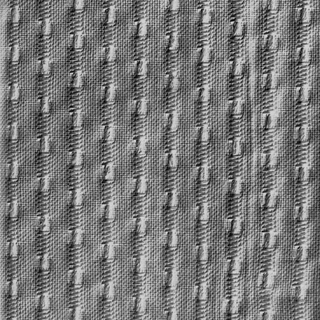}
    \end{subfigure}
    \begin{subfigure}[t]{0.095\textwidth}
        \includegraphics[width=\linewidth]{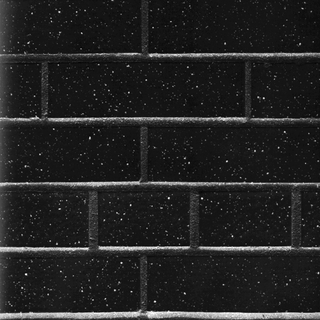}
    \end{subfigure}
    \begin{subfigure}[t]{0.095\textwidth}
        \includegraphics[width=\linewidth]{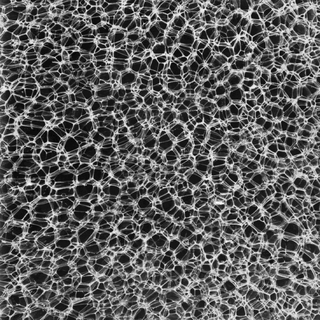}
    \end{subfigure}
    \begin{subfigure}[t]{0.095\textwidth}
        \includegraphics[width=\linewidth]{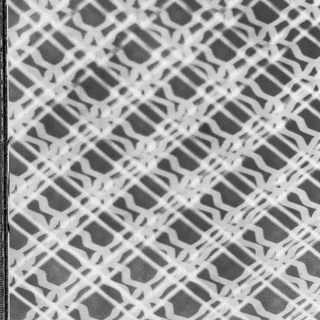}
    \end{subfigure}
    \begin{subfigure}[t]{0.095\textwidth}
        \includegraphics[width=\linewidth]{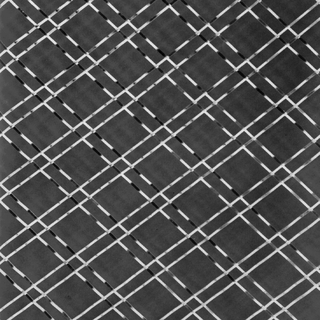}
    \end{subfigure}
    \begin{subfigure}[t]{0.095\textwidth}
        \includegraphics[width=\linewidth]{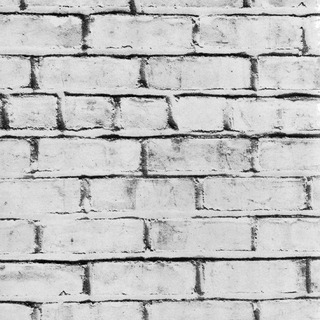}
    \end{subfigure}
    \begin{subfigure}[t]{0.095\textwidth}
        \includegraphics[width=\linewidth]{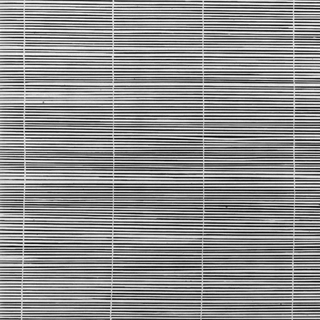}
    \end{subfigure}
    \begin{subfigure}[t]{0.095\textwidth}
        \includegraphics[width=\linewidth]{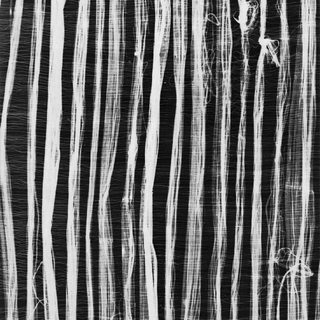}
    \end{subfigure}

    \caption{Brodatz Patterns used in the synthetic
      experiments}
    \label{fig:brodatz_patt}
\end{figure}

We compare our method to $\ncut$, using either graylevel intensities
or ``texture features'', where we use the magnitudes of the response
of 12 Gabor filters (3 wavelengths and 4 orientations) to define
appearance vectors for each pixel.

Figure \ref{fig:brodatz_examples_seg} shows some of the test images in
our dataset along with the computed eigenvectors and segmentations
arising with the proposed $\mixncut$ method and the different $\ncut$
formulations. The poor segmentation performance of $\ncut$ defined
over the raw pixel values (Figure \ref{fig:brodatz_examples_seg_ncut})
can be attributed to its inability to handle the complex appearance of
textured regions. This issue is partially solved when texture features
defined by Gabor filters are considered, but it has the drawback of
over-smoothing region boundaries (first and third examples in Figure
\ref{fig:brodatz_examples_seg_gabor}). In fact, in some extreme cases,
it misses an entire small region.  On the other hand, the new
$\mixncut$ method defined directly in terms of raw pixel values finds
near optimal segmentations in all of these examples, preserving well
the region boundaries and outperforming both baselines. This is due to
its capacity to model long range relationships without relying on
filtering methods.

We also compare our proposed algorithm to several state of art texture
segmentation methods. They include Level Set segmentation using
Wasserstein Distances (LSWD) \cite{ni2009local}, Factorization Based
Segmentation (FBS) \cite{yuan2015factorization}, Projective
Non-Negative Matrix Factorization on a Graph (PNMF)
\cite{bampis2016projective} and ORTSEG \cite{mccann2014images}. In all
these methods we try different combinations of parameters and select
the ones that performed the best in our data based on their $\jac$
value.  We run our comparison using the implementations provided by
their authors. FBS and PNMF use a bank of Gabor filters to define
texture features, whereas LSWD and ORTSEG use of local image
histograms.

Furthermore, we compare our method with the Multi-view Spectral
Clustering (MVSC) algorithm \cite{zhou2007spectral}.  MVSC uses a
convex combination of two Laplacian matrices representing different
graphs/views. For MVSC, we input the weight matrices of
$G_\text{grid}$ and $G_\text{data}$ to their algorithm and tested the
results using the same choices for $\lambda$ and $\sigma$ as in
$\mixcut$, picking the best segmentation according to the $\jac$
value.

For the quantitative experiments, we use all pairings of the 10
textures in Figure \ref{fig:brodatz_patt} with three different
ground-truth segmentations shown in
Table~\ref{tab:cut_results_brodatz} to generate three sets of images.
We compute the mean accuracy of each method on each set of images
using several parameter combinations.  Table
\ref{tab:cut_results_brodatz} summarizes the best mean accuracy
obtained with each method on each set of inputs. The table also shows
the average running time of each method.  We see the new $\mixncut$
approach obtains near perfect accuracy ($\jac \approx 1$) on all
ground-truth patterns, outperforming the other methods, specially the
spectral ones.


\begin{table}
\setlength{\fboxrule}{.5pt}
\setlength{\fboxsep}{0pt}
\centering
\caption{Evaluation of different segmentation methods on textured
  images.  The table summarizes accuracy and running time of each
  method on images with different ground-truth
  segmentations.}\label{tab:cut_results_brodatz}
\vspace{10pt}
\begin{adjustbox}{max width=\textwidth}
\begin{tabular}{ccccc}
&\multicolumn{3}{c}{\textbf{$\jac$ value}}& \\
\cmidrule(lr){2-4}
 \textbf{Method} & \includegraphics[width = .13\linewidth]{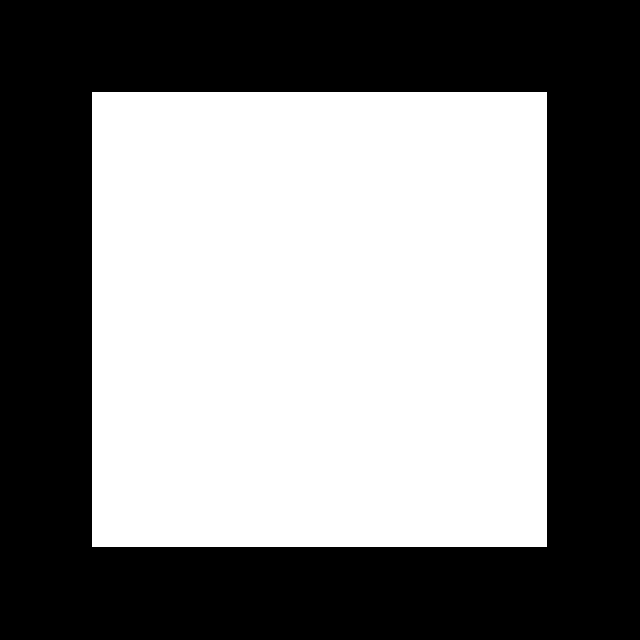} & \includegraphics[width = .13\linewidth]{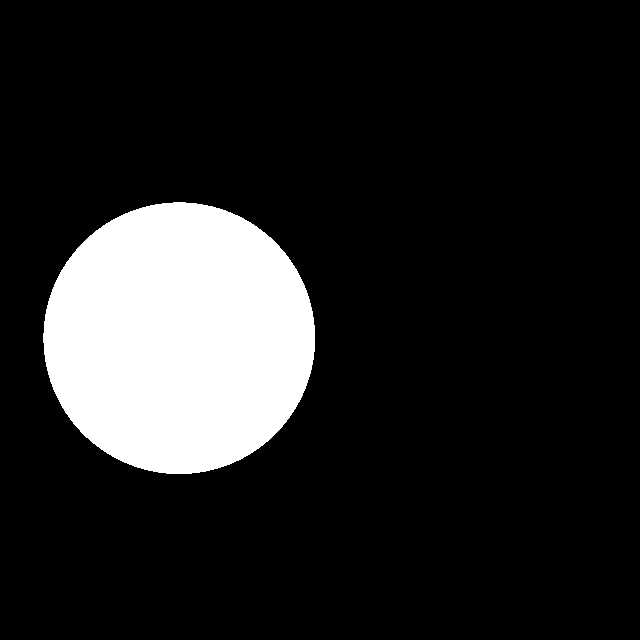}  & \includegraphics[width = .13\linewidth]{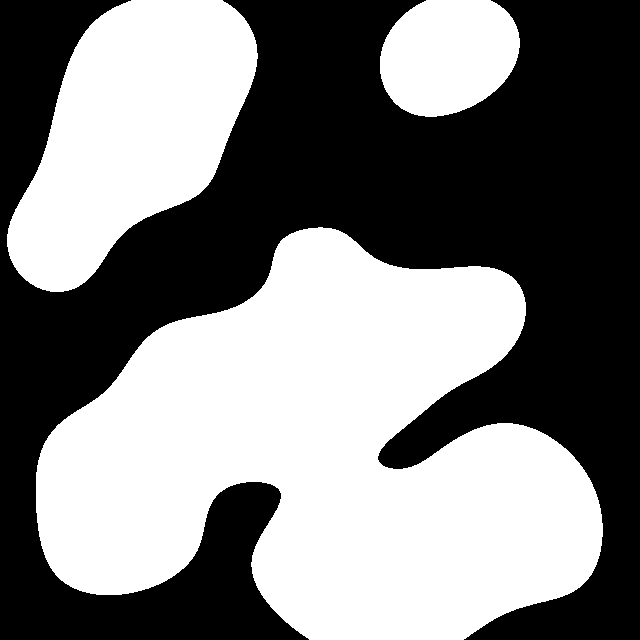}  & \textbf{Time (s)} \\\midrule[1pt]
 $\mixncut$
 & $0.906 \pm 0.080$  & $0.876 \pm 0.107$  & $0.842 \pm 0.133$  & 11.27  \\ 
   \midrule[1pt]
 $\ncut$-Graylevel
 & $0.541 \pm 0.128$  & $0.470 \pm 0.126$  & $0.538 \pm 0.130$  & 9.53 \\ 
$\ncut$-Gabor
 & $0.800 \pm 0.173$  & $0.779 \pm 0.197$  & $0.661 \pm 0.193$ & 13.14  \\

  MVSC
 & $0.731 \pm 0.273$  & $0.766 \pm 0.214$  & $0.656 \pm 0.284$ & 11.19  \\ 
LSWD
 & $0.852  \pm 0.129$ & $0.794  \pm 0.198$ & $0.828  \pm 0.119$ & 67.08  \\ 
ORTSEG
 & $0.853 \pm 0.149$ & $0.668 \pm 0.274$ & $0.826 \pm 0.135$ & 1.56 \\
FBS
 & $0.850 \pm 0.151$ & $0.778 \pm 0.214$ & $0.804 \pm 0.146$ & 0.10 \\
PNMF
  & $0.907  \pm 0.092$ & $0.733  \pm 0.142$ & $0.857  \pm 0.085$ & 14.59  \\ 
\bottomrule
\end{tabular}
\end{adjustbox}
\end{table}

\section{Conclusion}

We introduced a new spectral method for image segmentation that can
segment challenging images while working directly with raw pixel
values, without any pre-processing or filtering.  The approach is
based on a novel combination of appearance and spatial grouping cues
using two different graphs.  We use a dense graph to capture the
appearance of regions.  This leads to non-parametric models of region
appearance.  We also introduced a technique that can be used to
sparsify the resulting graph to ease the computational burden of
spectral segmentation.  Our results show that long range pairwise interactions
can capture the appearance of textured regions and significantly
improve the performance of graph-based segmentation methods.  The
proposed method is practical and it can be applied to
different types of images (natural scenes, biomedical, textures, etc.) that arise in a variety of application.

\section*{Broader Impact}

Image segmentation has a variety of applications that can benefit all
of society.  For example, segmentation methods may enable advances in
biomedical image analysis (including for medical diagnosis and
treatment), tele-conferencing technology, human-computer-interaction,
remote sensing, and robotics.  However, there are also potential uses
with questionable ethics, including mass surveillance and military
applications.

\bibliography{references}

\bibliographystyle{ieeetr}
\end{document}